\documentclass[12pt,dvips,generic,noinfoline]{imsart}

\RequirePackage[OT1]{fontenc}
\RequirePackage{amsthm,amsmath,amsfonts,amssymb,amsbsy,fancybox,natbib}
\RequirePackage[colorlinks,citecolor=blue,urlcolor=blue]{hyperref}
\RequirePackage{hypernat}
\usepackage[ruled,section]{algorithm}
\usepackage[noend]{algorithmic}
\usepackage{fullpage}
\usepackage{subfigure,epsfig,graphicx}
\usepackage{multirow,array,booktabs}

\startlocaldefs
\numberwithin{equation}{section}
\theoremstyle{plain}
\newtheorem{theorem}{Theorem}[section]

\newtheoremstyle{remark}{\topsep}{\topsep}%
     {\normalfont}
     {}           
     {\bfseries}  
     {.}          
     {.5em}       
     {\thmname{#1}\thmnumber{ #2}\thmnote{ #3}}
\theoremstyle{remark}

\newtheorem{assumption}{Assumption}[section]
\newtheorem{definition}{Definition}[section]
\endlocaldefs

\def\comma{\unskip,~}

\long\def\comment#1{}
\def\reals{{\mathbb R}}

\def\supp{\mathop{\text{supp}\kern.2ex}}
\def\argmin{\mathop{\text{\rm arg\,min}}}

\def\argmax{\mathop{\text{\rm arg\,max}}}
\let\hat\widehat

\let\hat\widehat

\def\given{{\,|\,}}
\def\ds{\displaystyle}

\def\1{{(1)}}
\def\2{{(2)}}

\def\tr{\mathop{\text{tr}}}
\long\def\comment#1{}
\def\True{\texttt{True}}
\def\False{\texttt{False}}

\begin{document}
\setcounter{page}{0}

\begin{frontmatter}
\title{Graph-Valued Regression}
\runtitle{Graph-Valued Regression}

\begin{aug}
\author{\fnms{Han} \snm{Liu}\ead[label=e1]{hanliu@cs.cmu.edu}}
\comma
\author{\;\fnms{Xi} \snm{Chen}\ead[label=e2]{xichen@cs.cmu.edu}}
\comma
\author{\;\fnms{John} \snm{Lafferty}\ead[label=e3]{lafferty@cs.cmu.edu}}
\and
\author{\fnms{Larry} \snm{Wasserman}\ead[label=e4]{larry@stat.cmu.edu}}

\address{Carnegie Mellon University\\[10pt]
}
     \end{aug}

\begin{abstract}
  Undirected graphical models encode in a graph $G$ the dependency
  structure of a random vector $Y$. In many applications, it is of
  interest to model $Y$ given another random vector $X$ as input.  We
  refer to the problem of estimating the graph $G(x)$ of $Y$
  conditioned on $X=x$ as ``graph-valued regression.''  In this paper,
  we propose a semiparametric method for estimating $G(x)$ that builds
  a tree on the $X$ space just as in CART (classification and
  regression trees), but at each leaf of the tree estimates a graph.
  We call the method ``Graph-optimized CART,'' or Go-CART. We study
  the theoretical properties of Go-CART using dyadic partitioning
  trees, establishing oracle inequalities on risk minimization and
  tree partition consistency. We also demonstrate the application of
  Go-CART to a meteorological dataset, showing how graph-valued
  regression can provide a useful tool for analyzing complex data.
\end{abstract}


\tableofcontents
\end{frontmatter}
\thispagestyle{empty}

\section{Introduction}

Let $Y$ be a $p$-dimensional random vector with distribution $P$.  A
common way to study the structure of $P$ is to construct the
undirected graph $G=(V,E)$, where the vertex set $V$ corresponds to
the $p$ components of the vector $Y$.  The edge set $E$ is a subset
of the pairs of vertices, where an edge between $Y_j$ and $Y_k$ is
absent if and only if $Y_j$ is conditionally independent of $Y_k$
given all the other variables. 
Suppose now that $Y$ and $X$ are both random vectors, and let
$P(\cdot \given X)$ denote the conditional distribution of $Y$ given
$X$.  In a typical regression  problem, we are interested in the
conditional mean $\mu(x) = \mathbb{E}\left( Y \given X=x \right)$.
But if $Y$ is multivariate, we may also be interested in how the
structure of $P(\cdot \given X)$ varies as a function of $X$.  In
particular, let $G(x)$ be the undirected graph corresponding to
$P(\cdot \given X=x)$. We refer to the problem of estimating $G(x)$
as {\em graph-valued regression}.

Let ${\cal G} = \{ G(x):\ x\in {\cal X}\}$ be a set of graphs
indexed by $x \in \mathcal{X}$, where ${\cal X}$ is the domain of
$X$.  Then ${\cal G}$ induces a partition of ${\cal X}$, denoted as
${\cal X}_1, \ldots, {\cal X}_m$, where $x_{1}$ and $x_{2}$ lie in
the same partition element if and only if $G(x_{1})=G(x_{2})$.
Graph-valued regression is thus the problem of estimating the partition and
estimating the graph within each partition element.

We present three different partition-based graph estimators; two
that use global optimization, and one based on a greedy splitting
procedure.  One of the optimization based schemes uses penalized
empirical risk minimization; the other uses held-out risk
minimization. As we show, both methods enjoy strong theoretical
properties under relatively weak assumptions; in particular, we
establish oracle inequalities on the excess risk of the estimators,
and tree partition consistency (under stronger assumptions) in
Section~\ref{sec::theory}. While the optimization based estimates
are attractive, they do not scale well computationally when the
input dimension is large.  An alternative is to adapt the greedy
algorithms of classical CART, as we describe in
Section~\ref{sec:greedyalgo}. In Section \ref{sec::experiment} we
present experimental results on both synthetic data and a
meteorological dataset, demonstrating how graph-valued regression
can be an effective tool for analyzing high dimensional data with
covariates.

\section{Graph-Valued Regression}

Let $y_1,\ldots, y_n$ be a random sample of vectors
from $P$, where each $y_i\in\reals^p$. We are interested in the case
where $p$ is large and, in fact, may diverge with $n$
asymptotically. One way to estimate $G$ from the sample is the {\em
graphical lasso} or {\em glasso} \citep{Yuan:Lin:07,
FHT:07,Banerjee:08}, where one assumes that $P$ is Gaussian with
mean $\mu$ and covariance matrix $\Sigma$. Missing edges in the
graph correspond to zero elements in the precision matrix $\Omega =
\Sigma^{-1}$ \citep{whit:1990, Edwa:1995, laur:1996}. A sparse
estimate of $\Omega$ is obtained by solving
\begin{eqnarray}
\hat{\Omega} = \argmin_{\Omega\succ 0 } \bigl\{{\rm tr}(S \Omega) -
\log |\Omega| + \lambda \|\Omega\|_1\bigr\} \label{eq::glasso}
\end{eqnarray}
where $\Omega$ is positive definite, $S$ is the sample covariance
matrix, and $\|\Omega\|_1 = \sum_{j,k} |\Omega_{jk}|$ is the
elementwise $\ell_{1}$-norm of $\Omega$. 
\Citet{FHT:07} develop a efficient algorithm for finding
$\hat\Omega$ that involves estimating a
single row (and column) of $\Omega$ in each iteration by solving a
lasso regression. The theoretical properties of $\hat\Omega$ have been
studied by \cite{Rothman:08} and \cite{Ravikumar:Gauss:09}. In
practice, it seems that the glasso yields reasonable graph estimators
even if $Y$ is not Gaussian; however, proving conditions under which
this happens is an open problem.

We briefly mention three different strategies for estimating $G(x)$,
the graph of $Y$ conditioned on $X=x$, each of which builds upon the
glasso.

\vskip5pt\noindent
{\bf Parametric Estimators.}  Assume that $Z=(X,Y)$ is jointly
multivariate Gaussian with covariance matrix
$$\Sigma = 
\left(
\begin{array}{cc}
  \Sigma_X & \Sigma_{XY}\\
  \Sigma_{YX} & \Sigma_{Y} \\
\end{array}
\right).$$ We can estimate
$\Sigma_{X}$, $\Sigma_{Y}$, and $\Sigma_{XY}$ by their corresponding
sample quantities $\hat{\Sigma}_{X}$, $\hat{\Sigma}_{Y}$, and
$\hat{\Sigma}_{XY}$, and the marginal precision matrix of $X$,
denoted as $\Omega_{X}$, can be estimated using the glasso.  The
conditional distribution of $Y$ given $X=x$ is obtained by standard
Gaussian formulas.  In particular, the conditional covariance matrix
of $Y\given X$ is $\hat{\Sigma}_{Y| X} = \hat{\Sigma}_{Y} -
\hat{\Sigma}_{YX}\hat{\Omega}_{X}\hat{\Sigma}_{XY}$ and a sparse
estimate of $\hat{\Omega}_{Y|X}$ can be
obtained by directly plugging $\hat{\Sigma}_{Y|X}$ into glasso.
However, the estimated graph does not vary with different values of
$X$.

\vskip5pt\noindent
{\bf Kernel Smoothing Estimators.} We assume that $Y$ given $X$ is
Gaussian, but without making any assumption about the marginal
distribution of $X$. Thus $Y \given X=x \sim N(\mu(x),\Sigma(x))$.
Under the assumption that both $\mu(x)$ and $\Sigma(x)$ are smooth
functions of $x$, we estimate $\Sigma(x)$ via kernel smoothing:
$$
\hat \Sigma(x) = \sum_{i=1}^n K\left(\frac{\|x-x_i\|}{h}\right)
     \left(y_i - \hat\mu(x)\right)\left(y_i - \hat\mu(x)\right)^T  \Bigm /
     {\sum_{i=1}^n K\left(\frac{\|x-x_i\|}{h}\right)}
$$
where $K$ is a kernel (e.g. the probability density function of the
standard Gaussian distribution), $\|\cdot\|$ is the Euclidean norm,
$h>0$ is a bandwidth and
$$
\hat\mu(x) = \sum_{i=1}^n K\left(\frac{\|x-x_i\|}{h}\right) y_i
     \Bigm / \sum_{i=1}^n K\left(\frac{\|x-x_i\|}{h}\right).
$$
Now we apply glasso  in \eqref{eq::glasso} with $S=\hat\Sigma(x)$ to
obtain an estimate of $ G(x)$.  This method is appealing because it
is simple and very similar to nonparametric regression smoothing;
the method was analyzed for one-dimensional $X$ by \cite{Zhou:10}.
However, while it is easy to estimate $G(x)$ at any given $x$, it
requires global smoothness of the mean and covariance functions. It is also computationally challenging to reconstruct the partition ${\cal X}_1, \ldots, {\cal X}_m$.

\vskip5pt\noindent
{\bf Partition Estimators.}  In this approach, we partition ${\cal
  X}$ into finitely many connected regions $\mathcal{X}_1, \ldots, \mathcal{X}_m$.  Within each
$\mathcal{X}_j$, we apply the glasso to get an estimated graph
$\widehat{G}_j$. We then take $\hat G(x) = \widehat{G}_j$ for all
$x\in \mathcal{X}_j$. To find the partition, we appeal to the idea
used in CART (classification and regression trees) \citep{cart:84}.
We take the partition elements to be recursively defined hyperrectangles.  As is well-known, we can then
represent the partition by a tree, where each leaf node corresponds
to a single partition element.  In CART, the leaves are associated
with the means within each partition element; while in our case, there will
be an estimated undirected graph for each leaf node. We refer to
this method as Graph-optimized CART, or Go-CART. The remainder of
this paper is devoted to the details of this method.

\section{Graph-Optimized CART}
 \label{sec::gocart}

 Let $X\in\reals^d$ and
$Y\in\reals^p$ be two random vectors, and let $\left\{ (x_{1},
y_{1}), \ldots, (x_{n}, y_{n}) \right\}$ be $n$ i.i.d.~samples from
the joint distribution of $(X,Y)$.
The domains of $X$ and $Y$ are denoted by $\mathcal{X}$ and
$\mathcal{Y}$ respectively; and for simplicity we take $\mathcal{X}
= [0,1]^{d}$.  We assume that
\begin{eqnarray*}
Y \given X = x \sim N_{p}(\mu(x), \Sigma(x))\label{eq.cgaussmodel}
\end{eqnarray*}
where $\mu: \mathbb{R}^{d} \rightarrow \mathbb{R}^{p}$ is a
vector-valued mean function and $\Sigma: \mathbb{R}^{d} \rightarrow
\mathbb{R}^{p\times p}$ is a matrix-valued covariance function.  We
also assume that for each $x$,  $\Omega(x) = \Sigma(x)^{-1}$ is a
sparse matrix, i.e., many elements of $\Omega(x)$ are zero.  In
addition, $\Omega(x)$ may also be a sparse function of $x$, i.e.,
$\Omega(x) = \Omega(x_{R})$ for some $R \subset \{1,\ldots, d\}$
with cardinality $|R| \ll d$. The task of graph-valued regression is
to find a sparse inverse covariance $\hat{\Omega}(x)$ to estimate
$\Omega(x)$ for any $x \in \mathcal{X}$; in some situations the
graph of $\Omega(x)$ is of greater interest than the entries of
$\Omega(x)$ themselves.

Go-CART is a partition-based conditional graph estimator. We
partition ${\cal X}$ into finitely many connected regions
$\mathcal{X}_1, \ldots, \mathcal{X}_m$, and within each
$\mathcal{X}_j$ we apply the glasso to estimate a graph
$\hat{G}_j$. We then take $\hat G(x) = \hat{G}_j$ for all $x\in
\mathcal{X}_j$. To find the partition,  we restrict ourselves to
dyadic splits, as studied by \cite{scott06} and \cite{blanchard07}.  The
primary reason for such a  choice is the computational and
theoretical tractability of dyadic partition-based estimators.

\subsection{Dyadic Partitioning Tree}

Let $\mathcal{T}$ denote the set of dyadic partitioning trees (DPTs)
defined over $\mathcal{X}=[0,1]^{d}$, where each DPT $T \in
\mathcal{T}$ is constructed by recursively dividing $\mathcal{X}$ by
means of axis-orthogonal dyadic splits.  Each node of a DPT
corresponds to a hyperrectangle in $[0,1]^{d}$. If a node is
associated to the hyperrectangle $\mathcal{A}=\prod_{l=1}^d [a_l,
b_l]$, then after being dyadically split along dimension $k$, the
two children are associated with the sub-hyperrectangles
\begin{equation*}
\mathcal{A}^{(k)}_L=\prod_{l<k} [a_l, b_l] \times [a_k,
\frac{a_k+b_k}{2}] \times \prod_{l>k} [a_l, b_l]~~\text{and}~~\mathcal{A}^{(k)}_R= \mathcal{A} \backslash \mathcal{A}^{(k)}_L.
\end{equation*}
Given a DPT $T$, we denote by $\Pi(T) = \{\mathcal{X}_{1}, \ldots,
\mathcal{X}_{m_{T}} \}$ the partition of $\mathcal{X}$ induced by
the leaf nodes of $T$.
For a dyadic integer $N=2^{K}$ where $K\in\{0,1, 2,  \ldots \}$, we define $\mathcal{T}_{N}$ to be
the collection of all DPTs such that no partition has a side length
smaller than $2^{-K}$.
Let $I(\cdot)$ denote the indicator function. We denote $\mu_{T}(x)$  and $\Omega_{T}(x)$ as the piecewise
constant mean and precision functions associated with $T$:
\begin{eqnarray*}
\mu_{T}(x)  = \sum_{j=1}^{m_{T}}\mu_{\mathcal{X}_j} \cdot I \left( x
\in \mathcal{X}_{j} \right)~~\text{and}~~ \Omega_{T}(x)  =
\sum_{j=1}^{m_{T}}\Omega_{\mathcal{X}_{j}} \cdot I\left( x \in
\mathcal{X}_{j}\right),
\end{eqnarray*}
where  $\mu_{\mathcal{X}_{j}} \in \mathbb{R}^{p}$ and
$\Omega_{\mathcal{X}_{j}}\in\mathbb{R}^{p\times p}$ are the mean
vector and precision matrix for $\mathcal{X}_j$.  

\subsection{Go-CART: Risk Minimization Estimator}

Before formally defining our graph-valued regression estimators, we
require some further definitions.  Given a DPT $T$ with an induced
partition $\Pi(T) = \{\mathcal{X}_{j}\}_{j=1}^{m_{T}}$ and
corresponding mean and precision functions $\mu_{T}(x)$ and
$\Omega_{T}(x)$, the negative conditional log-likelihood risk $R(T,
\mu_{T}, \Omega_{T})$ and its sample version $\hat{R}(T, \mu_{T},
\Omega_{T})$ are defined as follows:
\begin{eqnarray}
R(T, \mu_{T}, \Omega_{T}) =  \sum_{j=1}^{m_{T}}\mathbb{E} \biggl[
\Bigl( \mathrm{tr}\Bigl[ \Omega_{\mathcal{X}_{j}} \left(
(Y-\mu_{\mathcal{X}_{j}}) (Y-\mu_{\mathcal{X}_{j}})^{T} \right)
\Bigr]  -
\log |\Omega_{\mathcal{X}_{j}}|\Bigr)\cdot I\left(X \in \mathcal{X}_{j}\right) \biggr], \\
\hat{R}(T, \mu_{T}, \Omega_{T}) =
\frac{1}{n}\sum_{i=1}^{n}\sum_{j=1}^{m_{T}} \biggl[\Bigl(
\mathrm{tr} \Bigl[ \Omega_{\mathcal{X}_{j}}\left(
(y_{i}-\mu_{\mathcal{X}_j}) (y_{i}-\mu_{\mathcal{X}_{j}})^{T}
\right) \Bigr]- \log |\Omega_{\mathcal{X}_{j}}|\Bigr)\cdot I \left(
x_{i} \in \mathcal{X}_{j}\right)\biggr].  \label{eq::erisk}
\end{eqnarray}
Let $[[T]]>0$ denote a prefix code over all DPTs $T \in
\mathcal{T}_{N}$ satisfying $\sum_{T \in \mathcal{T}_{N}}2^{-[[T]] }
\leq 1$.  One such prefix code $[[T]]$ is proposed in
\citep{scott06}, and takes the form $$ [[T]] =3|\Pi(T)| - 1
+(|\Pi(T)|-1)\log d/\log 2.$$  A simple upper bound for $[[T]]$ is
\begin{equation}
[[T]]\leq (3 + \log d/\log2) |\Pi(T)|. \label{eq.prefixcode}
\end{equation}
Our analysis will assume that the conditional means and precision
matrices are bounded in the $\|\cdot\|_\infty$ and $\|\cdot\|_1$
norms; specifically we suppose there is a positive constant $B$ and
a sequence $L_{1,n}, \ldots, L_{m_{T}, n} $, where each $L_{j,n}
\in \mathbb{R}^{+}$ is a function of the sample size $n$, and we
define the domains of each $\mu_{\mathcal{X}_{j}}$ and
$\Omega_{\mathcal{X}_j}$ as
\begin{eqnarray}
M_{j} & = & \left\{ \mu \in \mathbb{R}^{p}:\ \|\mu \|_{\infty} \leq
  B\right\}, \nonumber \\
\Lambda_{j} & = & \left\{\Omega \in \mathbb{R}^{p\times p}:\ \Omega
~\text{is positive definite, symmetric, and } \|\Omega \|_{1} \leq
L_{j,n}  \right\}. \label{eq::Lambda}
\end{eqnarray}

With this notation in place, we can now define two estimators.
\begin{definition}
\label{def:risk} \enspace The {\it penalized empirical risk
minimization Go-CART estimator} is defined as
\begin{eqnarray*}
\hat{T}, \left\{\hat{\mu}_{\hat{\mathcal{X}}_{j}},
\hat{\Omega}_{\hat{\mathcal{X}}_{j}}\right\}_{j=1}^{m_{\hat{T}}} =
\ds {\rm argmin}_{T\in  \mathcal{T}_{N}, \mu_{\mathcal{X}_j}\in
M_{j}, \Omega_{\mathcal{X}_j} \in \Lambda_{j}}\biggl\{ \hat{R}(T,
\mu_{T}, \Omega_{T}) + \mathrm{pen}(T)\biggr\}
\end{eqnarray*}
where $\hat{R}$ is defined in \eqref{eq::erisk} and $$\mathrm{pen}(T)
= \gamma_{n}\cdot m_{T}\sqrt{\frac{[[T]] \log 2 + 2\log (np)}{n}}.$$
\end{definition}
Empirically, we may always set the dyadic integer $N$ to be a
reasonably large value; the regularization parameter $\gamma_{n}$ is
responsible for selecting a suitable DPT $T \in \mathcal{T}_{N}$. Once
$T$ is chosen, the tuning parameters $L_{1,n}, \ldots, L_{m_{T},
  n}$ corresponding each partition element of $T$ need to be
determined in a data-dependent way.  We will discuss further details
about this in the next section.

We can also formulate an estimator by minimizing held-out
risk. Practically, we split the data into two partitions;
we use $\mathcal{D}_{1} =\{(x_{1}, y_{1}), \ldots, (x_{n_{1}}, y_{n_{1}})\}$
for training and $\mathcal{D}_{2} =\{((x'_{1}, y'_{1}), \ldots,
(x'_{n_{2}}, y'_{n_{2}}))\}$ for validation with $n_{1} + n_{2} =n$.
The held-out negative log-likelihood risk is then given by
\begin{align}
\nonumber
\lefteqn{ \hat{R}_{\rm out}(T, \mu_{T}, \Omega_{T}) =} ~~~~~& \\
&\quad \frac{1}{n_{2}}\sum_{i=1}^{n_{2}}\sum_{j=1}^{m_{T}}\Bigl\{
\Bigl(\mathrm{tr}\Bigl[ \Omega_{\mathcal{X}_{j}}\left(
(y'_{i}-\mu_{\mathcal{X}_{j}}) (y'_{i}-\mu_{\mathcal{X}_{j}})^{T}
\right) \Bigr]
 - \log |\Omega_{\mathcal{X}_{j}}|\Bigr)\cdot I\left(x'_{i} \in \mathcal{X}_{j} \right)\Bigr\}. \label{eq::heldoutrisk}
\end{align}

\begin{definition}
 \label{def:ho}
\enspace For each DPT $T$ define
\begin{eqnarray}
\hat{\mu}_{T}, \hat{\Omega}_{T} = \ds {\rm
argmin}_{\mu_{\mathcal{X}_j}\in M_{j}, \Omega_{\mathcal{X}_j} \in
\Lambda_{j}}\hat{R}(T, \mu_{T}, \Omega_{T})  \label{eq::step1est}
\end{eqnarray}
where $\hat{R}$ is defined in \eqref{eq::erisk} but only evaluated
on $\mathcal{D}_{1} =\{(x_{1}, y_{1}), \ldots,
(x_{n_{1}},y_{n_{1}})\}$. The {\it held-out risk minimization
Go-CART estimator} is
\begin{eqnarray*}
\hat{T}= \ds {\rm argmin}_{T\in  \mathcal{T}_{N}}\hat{R}_{\rm
out}(T, \hat{\mu}_{T}, \hat{\Omega}_{T}) .
\end{eqnarray*}
where $\hat{R}_{\rm out}$ is defined in \eqref{eq::heldoutrisk} but only evaluated on $\mathcal{D}_{2}$.
\end{definition}

\subsection{Go-CART: Greedy Partitioning}

\label{sec:greedyalgo}

The above procedures require us to find an
optimal dyadic partitioning tree within $\mathcal{T}_{N}$.  Although
dynamic programming can be applied, as in \citep{blanchard07},
the computation does not scale to large input dimensions $d$.  We
now propose a simple yet effective greedy algorithm to
find an approximate solution $(\hat{T}, \hat{\mu}_{T},
\hat{\Omega}_{T})$.  We focus on the 
held-out risk minimization form as in Definition \ref{def:ho},
due to its superior empirical  performance. But note that our greedy
approach is generic and can easily be adapted to the 
penalized empirical risk minimization form.  

First, consider the simple case that we are given a dyadic tree
structure $T$ which induces a partition $\Pi(T)\!\!=\!\!\{\mathcal{X}_{1},
\ldots, \mathcal{X}_{m_{T}}\}$ on $\mathcal{X}$.  For any partition
element $\mathcal{X}_{j}$, we estimate the sample mean using $\mathcal{D}_{1}$:
\begin{equation*}
  \hat{\mu}_{\mathcal{X}_{j}}= \frac{1}{\sum_{i=1}^{n_{1}} I\left( x_{i} \in \mathcal{X}_{j}\right)}\sum_{i=1}^{n_{1}} y_{i} \cdot I\left( x_{i} \in \mathcal{X}_{j}\right).
\end{equation*}
The glasso is then used to estimate a sparse precision
matrix $\hat{\Omega}_{\mathcal{X}_{j}}$.  More precisely, let
$\hat{\Sigma}_{\mathcal{X}_{j}}$ be the sample covariance matrix
for the partition element $\mathcal{X}_{j}$, given by
\begin{equation*}
\hat{\Sigma}_{\mathcal{X}_{j}}= \frac{1}{ \sum_{i=1}^{n_{1}}I \left( x_{i} \in \mathcal{X}_{j}\right)}
\sum_{i=1}^{n_{1}} \left( y_{i}-\hat{\mu}_{\mathcal{X}_{j}} \right)
\left(y_{i}-\hat{\mu}_{\mathcal{X}_{j}} \right)^T\cdot I\left(x_{i} \in \mathcal{X}_{j} \right).
\end{equation*}
The estimator  $\hat{\Omega}_{\mathcal{X}_{j}}$ is obtained by
optimizing $$
 \hat{\Omega}_{\mathcal{X}_{j}} = \argmin_{\Omega \succ 0}\{ \tr(
 \hat{\Sigma}_{\mathcal{X}_{j}}\Omega) -\log|\Omega| +\lambda_{j}
 \|\Omega\|_1 \},$$
where $\lambda_{j}$  is in one-to-one correspondence with $L_{j,n}$
in \eqref{eq::Lambda}.  In practice, we run the full regularization
path of the glasso, from large $\lambda_{j}$, which yields very sparse
graph, to small $\lambda_{j}$, and select the graph that minimizes the
held-out negative log-likelihood risk.  
To further improve the model selection
performance, we refit the parameters of the precision matrix
after the graph has been selected.
That is, to reduce the bias of the glasso,
we first estimate the
sparse precision matrix using $\ell_1$-regularization,
and then we refit the Gaussian model without $\ell_1$-regularization, but
enforcing the sparsity pattern obtained in the first step.
\cite{Liu:10} demonstrate that such a refitting step will yield a
significantly better model selection performance when estimating graphs.

The natural, standard greedy procedure
starts from the coarsest
partition $\mathcal{X}=[0,1]^{d}$ and then computes the decrease in
the held-out risk by  dyadically splitting each hyperrectangle
$\mathcal{A}$ along dimension $k \in \{1,\ldots d\}$.
The dimension $k^{\ast}$ that results in the largest decrease
in held-out risk is selected. More precisely, let $\text{sl}_{k}(\mathcal{A})$ be
the side length of $\mathcal{A}$ on the dimension $k$. If
$\text{sl}_{k}(\mathcal{A})
> 2^{-K}$, where $K = \log_{2}N$, we dyadically split $\mathcal{A}$ along the dimension $k$.
In this case, let $\mathcal{A}_{L}^{(k)}$ and $\mathcal{A}_R^{(k)}$
be the two resulting sub-hyperrectangles. The decrease in held-out risk
takes the form
\begin{align}
\label{eq:DR}
 \Delta  \widehat{R}^{(k)}_{\mathrm{out}}(\mathcal{A}, \hat{\mu}_\mathcal{A}, \widehat{\Omega}_\mathcal{A}) 
 & = &     \hat{R}_{\mathrm{out}}(\mathcal{A}, \widehat{\mu}_\mathcal{A}, \widehat{\Omega}_\mathcal{A}) -
 \hat{R}_{\mathrm{out}}(\mathcal{A}^{(k)}_{L}, \hat{\mu}_{\mathcal{A}^{(k)}_{L}}, \widehat{\Omega}_{\mathcal{A}^{(k)}_{L}}) -
   \hat{R}_{\mathrm{out}}(\mathcal{A}^{(k)}_{R}, \hat{\mu}_{\mathcal{A}^{(k)}_{R}}, \widehat{\Omega}_{\mathcal{A}^{(k)}_{R}}).
\end{align}
Note that if splitting any dimension $k$ of $\mathcal{A}$ leads to
an increase in the risk, we set a Boolean variable
$S(\mathcal{A})=\False$ which indicates that the partition
element $\mathcal{A}$ should no longer be split and hence
$\mathcal{A}$ should be a partition element of $\Pi(T)$. The greedy
Go-CART, as presented in Algorithm \ref{algo:treegrow}, recursively
applies the previous procedure to split each partition element until
all the partition elements cannot be further split. Note that we
also record the dyadic partition tree structure in the
implementation.

\def\stateskip{\vskip 2pt}
\begin{algorithm}[!ht]
\begin{small}
\caption{Greedy Go-CART using Dyadic Partitioning}

\vskip10pt
\begin{algorithmic}
\STATE \textbf{Input:} training data $\{x_i, y_i\}_{i=1}^{n_{1}}$,
held-out validation data $\{x'_i, y'_i\}_{i=1}^{n_{2}}$, and 
an integer $K$

\stateskip

  \STATE Start from $\mathcal{X}=[0,1]^d$. Set the Boolean
  variable
  $S(\mathcal{X})=\True$ and estimate $\widehat{\mu}_\mathcal{X},
  \widehat{\Omega}_\mathcal{X}$

\stateskip

  \WHILE {exists a hyperrectangle $\mathcal{A}$ such that   $S(\mathcal{A})=\True$}
        \FORALL{ dimensions $k \in \{1,\ldots d\}$ }
            \IF {$\text{sl}_k(\mathcal{A}) \geq  2^{-K+1}$}
               \STATE Calculate $\Delta  \widehat{R}^{(k)}_{\mathrm{out}}(\mathcal{A}, \hat{\mu}_\mathcal{A}, \widehat{\Omega}_\mathcal{A})$  according to  \eqref{eq:DR}
            \ELSE
               \STATE Set  $\Delta  \widehat{R}^{(k)}_{\mathrm{out}}(\mathcal{A}, \hat{\mu}_\mathcal{A},
               \widehat{\Omega}_\mathcal{A})= - \infty$
            \ENDIF
        \ENDFOR
       \STATE Determine the best splitting dimension
       $k^{\ast}= \argmax_{k \in \{1,\ldots, d\}} \Delta  \widehat{R}^{(k)}_{\mathrm{out}}(\mathcal{A}, \hat{\mu}_\mathcal{A}, \widehat{\Omega}_\mathcal{A})$
      \IF {$\Delta  \widehat{R}^{(k^{\ast})}_{\mathrm{out}}(\mathcal{A}, \hat{\mu}_\mathcal{A}, \widehat{\Omega}_\mathcal{A})>0$}
       \STATE{Dyadically split  $\mathcal{A}$  along dimension
         $k^{\ast}$, yielding two hyperrectangles $\mathcal{A}_{L}^{(k^{\ast})}$ and $\mathcal{A}_{R}^{(k^{\ast})}$.
       Estimate $\widehat{\mu}_{\mathcal{A}_L^{(k^{\ast})}},
       \widehat{\Omega}_{\mathcal{A}_L^{(k^{\ast})}}, \widehat{\mu}_{\mathcal{A}_R^{(k^{\ast})}},
       \widehat{\Omega}_{\mathcal{A}_R^{(k^{\ast})}}$}  and set
       $S(\mathcal{A}_L^{(k^{\ast})})=S(\mathcal{A}_R^{(k^{\ast})})=\True$.
   \ELSE
       \STATE Set $S(\mathcal{A})=\False$ and put
       $\mathcal{A}$ into the final partition set.
   \ENDIF
  \ENDWHILE
\stateskip
\textbf{Output:} Partition
$\Pi(\hat{T})=\{\mathcal{X}_j\}_{j=1}^{m_{\hat{T}}}$ 
and the corresponding DPT $\hat{T}$ with the estimated $\widehat{\mu}_{\mathcal{X}_j}$.
\end{algorithmic}
\label{algo:treegrow}
\end{small}
\end{algorithm}

This greedy partitioning method parallels the classical algorithms for
classification and regression trees that have been used in statistical
learning for decades.  However, the strength of the procedures given
in Definitions~\ref{def:risk} and~\ref{def:ho} is that they lend
themselves to a theoretical analysis under relatively weak
assumptions, as we show in the following section. The theoretical
properties of greedy Go-CART are left to future work.

\section{Theoretical Properties}
\label{sec::theory}

We define the oracle risk $R^{*}$ over $\mathcal{T}_{N}$ as
\begin{equation*}
R^{*} =  R(T^{*}, \mu^{*}_{T}, \Omega^{*}_{T}) = \inf_{T\in
\mathcal{T}_{N}, \mu_{\mathcal{X}_j}\in M_j,
  \Omega_{\mathcal{X}_j} \in \Lambda_{j}}R(T, \mu_{T}, \Omega_{T}).
\end{equation*}
Note that $T^{*}$, $\mu^{*}_{T^{*}}$, and $\Omega^{*}_{T^{*}}$ might
not be unique, since the finest partition always achieves the oracle
risk.  To obtain oracle inequalities, we
make the following two technical assumptions.

\begin{assumption} \label{assump1} Let $T\in \mathcal{T}_{N}$ be an
  arbitrary DPT which induces a partition $\Pi(T) = \{\mathcal{X}_{1}, \ldots,
  \mathcal{X}_{m_{T}}\}$ on $\mathcal{X}$, we assume that there exists a
  constant $B$, such that
\begin{equation*}
\max_{1\leq j\leq m_{T}} \| \mu_{\mathcal{X}_{j}}\|_{\infty} \leq B
~~~~\mathrm{and}~~\max_{1\leq j\leq m_{T}}\sup_{\Omega \in
\Lambda_{j}} \log |\Omega|  \leq L_{n} \label{eq::sigmaeigen}
\end{equation*}
where $\Lambda_{j}$ is defined in \eqref{eq::Lambda} and $L_{n} =
\max_{1\leq j \leq m_{T}} L_{j,n}$, where $L_{j,n}$ is the same as
in \eqref{eq::Lambda}. We also assume that
\begin{equation*}
L_{n} = o(\sqrt{n}).
\end{equation*}
\end{assumption}

\vskip5pt
\begin{assumption}\label{assump2}
Let $Y = (Y_{1}, \ldots, Y_{p})^{T} \in \mathbb{R}^{p}$. For any   $\mathcal{A} \subset \mathcal{X}$, we define
\begin{eqnarray*}
Z_{k\ell}(\mathcal{A})  & = &Y_{k}Y_{\ell}\cdot I(X\in \mathcal{A}) - \mathbb{E}(Y_{k}Y_{\ell} \cdot I(X\in \mathcal{A})) \\
Z_{j}(\mathcal{A}) & = & Y_{j} \cdot I(X\in \mathcal{A})- \mathbb{E}(Y_{j}\cdot I(X\in \mathcal{A})).
\end{eqnarray*}
We assume  there exist constants $M_{1}, M_{2}, v_{1},$ and $v_{2}$,
such that
\begin{equation*}
\sup_{k, \ell, \mathcal{A}} \mathbb{E}|Z_{k\ell}(\mathcal{A})|^{m} \leq \frac{m!
M_{2}^{m-2}v_{2}}{2}~~\mathrm{and}~~\sup_{j,
\mathcal{A}}\mathbb{E}|Z_{j}(\mathcal{A})|^{m}\leq \frac{m! M_{1}^{m-2}v_{1}}{2}
\end{equation*}
for all $m\geq 2$.
\end{assumption}

\begin{theorem}\label{thm.oracle}
  Let $T\in \mathcal{T}_{N}$ be a DPT that induces a partition
  $\Pi(T) = \{\mathcal{X}_{1}, \ldots, \mathcal{X}_{m_{T}}\}$ on $\mathcal{X}$.
  For any $\delta \in (0,1)$, let $\hat{T},
  \hat{\mu}_{\hat{T}}, \hat{\Omega}_{\hat{T}}$ be the estimator
  obtained using the penalized empirical risk minimization Go-CART
  in Definition~\ref{def:risk}, with a penalty term $\mathrm{pen}(T)$
  of the form
\begin{eqnarray*}
\mathrm{pen}(T) = (C_{1}+1) L_{n}m_{T} \sqrt{\frac{[[T]]\log 2
+2\log p +\log(48/\delta)}{n}}
\end{eqnarray*}
where $C_{1} = 8 \sqrt{v_{2}} + 8 B \sqrt{v_{1}}+ B^{2}$. Then for
sufficiently large $n$, the excess risk inequality
\begin{eqnarray*}
R(\hat{T},\hat{\mu}_{\hat{T}}, \hat{\Omega}_{\hat{T}}) - R^{*} \leq
\inf_{T \in \mathcal{T}_{N}}\left\{  2 \mathrm{pen}(T) +
\inf_{\mu_{\mathcal{X}_j}\in M_{j}, \Omega_{\mathcal{X}_j} \in
\Lambda_{j}}(R(T, \mu_{T}, \Omega_{T}) - R^{*}) \right\}
\end{eqnarray*}
holds with probability at least $1-\delta$.
\end{theorem}

A similar oracle inequality holds when using the held-out risk
minimization Go-CART.

\begin{theorem}\label{thm.oracle2}
Let $T\in \mathcal{T}_{N}$ be a DPT which induces a partition
$\Pi(T)=\{\mathcal{X}_{1}, \ldots, \mathcal{X}_{m_{T}}\}$ on $\mathcal{X}$.  For any $\delta\in(0,1)$, we
define $\phi_{n}(T)$ to be a function of $n$ and $T$:
\begin{equation*}
\phi_{n}(T) =  (C_{2}+\sqrt{2}) L_{n}m_{T}
\sqrt{\frac{[[T]]\log 2 +2\log p +\log(384/\delta)}{n}}
\end{equation*}
where $C_{2} = 8 \sqrt{2v_{2}} + 8 B \sqrt{2v_{1}}+ \sqrt{2}B^{2}$
and $L_{n} = \max_{1\leq j \leq m_{T}} L_{j,n}$.  Partition the data
into $\mathcal{D}_{1} =\{(x_{1}, y_{1}), \ldots, (x_{n_{1}},
y_{n_{1}})\}$ and $\mathcal{D}_{2} =\{(x'_{1}, y'_{1}), \ldots,
(x'_{n_{2}}, y'_{n_{2}})\}$ with sizes $n_{1} = n_{2} = n/2$. Let
$\hat{T}, \hat{\mu}_{\hat{T}}, \hat{\Omega}_{\hat{T}}$ be the
estimator constructed using the held-out risk minimization criterion
of Definition~\ref{def:ho}. Then, for sufficiently large $n$, the
excess risk inequality
\begin{eqnarray*}
R(\hat{T},\hat{\mu}_{\hat{T}}, \hat{\Omega}_{\hat{T}}) - R^{*} \leq
\inf_{T \in \mathcal{T}_{N}}\left\{  3 \phi_{n}(T) +
\inf_{\mu_{\mathcal{X}_j}\in M_{j}, \Omega_{\mathcal{X}_j} \in
\Lambda_{j}}(R(T, \mu_{T}, \Omega_{T}) - R^{*}) \right\} +
\phi_{n}(\hat{T})
\end{eqnarray*}
holds with probability at least $1-\delta$.
\end{theorem}

Note that in contrast to the statement in Theorem~\ref{thm.oracle},
Theorem~\ref{thm.oracle2} results in a stochastic upper bound due to
the extra $\phi_{n}(\hat{T})$ term, which depends on the complexity
of the final estimate $\hat{T}$.  The 
proofs of both theorems are given in the appendix.


We now temporarily make the strong assumption that the model is
correct, so that $Y$ given $X$ is conditionally Gaussian, with a
partition structure that is given by a dyadic tree.  We show that
with high probability, the true dyadic partition structure can be
correctly recovered.

\begin{assumption}\label{assump3}
The true model is
\begin{eqnarray}
Y \given X = x \sim N_{p}(\mu^{*}_{T^{*}}(x),
\Omega^{*}_{T^{*}}(x))\label{eq.truemodel}
\end{eqnarray}
where $T^{*} \in \mathcal{T}_{N}$ is a DPT with induced partition
$\Pi(T^{*}) = \{\mathcal{X}^{0}_{j}\}_{j=1}^{m_{T^{*}}}$ and
\begin{equation*}
\mu^{*}_{T^{*}}(x) = \sum_{j=1}^{m_{T^{*}}} \mu^{*}_{j} \,I(x
\in \mathcal{X}^{0}_{j}),  ~~~~      \Omega^{*}_{T^{*}}(x) = \sum_{j=1}^{m_{T^{*}}} \Omega^{*}_{j} \,I(x
\in \mathcal{X}^{0}_{j}).
\end{equation*}
\end{assumption}
Under this assumption, clearly
\begin{eqnarray*}
R(T^{*}, \mu^{*}_{T^{*}}, \Omega^{*}_{T^{*}}) =
\inf_{T\in  \mathcal{T}_{N}, \mu_{T}, \Omega_{T}\in
\mathcal{M}_{T}}R(T, \mu_{T}, \Omega_{T}),
\end{eqnarray*}
where  $\mathcal{M}_{T}$ is given by
\begin{eqnarray}
\lefteqn{\mathcal{M}_{T} = \Bigl\{ \mu(x) =
  \sum_{j=1}^{m_{T}}\mu_{\mathcal{X}_{j}} \,I(x \in \mathcal{X}_{j}), \;
\Omega(x) = \sum_{j=1}^{m_{T}}\Omega_{\mathcal{X}_{j}}
\,I(x\in\mathcal{X}_{j}) :\ }~~~~~~~~~~~~~~~~~~~~~~~~~~~~~~~~~~~~ \nonumber\\
& & \text{where}~~~  \mu_{\mathcal{X}_{j}} \in M_{j}, \;
\Omega_{\mathcal{X}_{j}}\in \Lambda_{j}, \; \Pi(T) = \{\mathcal{X}_{j}\}_{j=1}^{m_{T}}\Bigr\}. \nonumber
\end{eqnarray}

Let $T_{1}$ and $T_{2}$ be two DPTs, if $\Pi({T_{1}})$ can be obtained by further split the hyperrectangles within $\Pi({T_{2}})$,  we  say $\Pi(T_{2}) \subset \Pi(T_{1})$. We then have the following definitions:
\begin{definition}
A tree estimation procedure $\hat T$ is  {\it tree partition
consistent} in case
\begin{eqnarray*}
\mathbb{P}\left(  \Pi(T^{*}) \subset \Pi(\hat{T})\right) \rightarrow
1~~\text{as}~n\rightarrow \infty.
\end{eqnarray*}
\end{definition}

Note that the estimated partition may be finer than the true
partition. Establishing a tree partition consistency result requires
further technical assumptions.  The following assumption specifies
that for arbitrary adjacent subregions of the true dyadic partition,
either the means or the variances should be sufficiently different.
Without such an assumption, of course, it is impossible to detect
the boundaries of the true partition.

\begin{assumption}\label{assump4}
  Let $\mathcal{X}^{0}_i$ and
  $\mathcal{X}^{0}_j$ be adjacent partition elements of $T^*$, so that
  they have a common parent node within $T^{*}$. Let $\Sigma^{*}_{ \mathcal{X}^{0}_i} =(
  \Omega^{*}_{\mathcal{X}^{0}_i})^{-1}$.  We assume there exist
  positive constants $ c_{1}, c_{2}, c_{3}, c_{4}$, such that either
\begin{eqnarray*}
2\log\left|\frac{\Sigma^*_{\mathcal{X}^{0}_i} +
    \Sigma^*_{\mathcal{X}^{0}_j} }{2}\right| -
\log|\Sigma^*_{\mathcal{X}^{0}_i}| -
\log|\Sigma^*_{\mathcal{X}^{0}_j}| \geq c_{4}
\end{eqnarray*}
or $\|\mu^{*}_{\mathcal{X}^{0}_i} - \mu^{*}_{\mathcal{X}^{0}_j}
\|^{2}_{2} \geq c_{3}$. We also assume
\begin{eqnarray*}
\rho_{\rm \min}(\Omega^{*}_{\mathcal{X}^{0}_{j}}) \geq c_{1}, \quad
\forall j = 1, \ldots, m_{T^{*}},
\end{eqnarray*}
where $\rho_{\min}(\cdot)$ denotes the smallest eigenvalue.
Furthermore, for any $T \in \mathcal{T}_{N}$ and any $\mathcal{A}
\in \Pi(T)$, we have $\mathbb{P}\left( X\in \mathcal{A}  \right)
\geq c_{2}$.
\end{assumption}

\vskip10pt

\begin{theorem}\label{thm.sparsistent} Under the above assumptions, we have
\begin{equation*}
\inf_{T\in \mathcal{T}_{N}, \;\Pi(T^{*}) \nsubseteq \Pi(T)}
\;\inf_{\mu_{T}, \,\Omega_{T}\in\mathcal{M}_{T}}R(T, \mu_{T},
\Omega_{T}) - R(T^{*},
\mu^*_{T^{*}}, \Omega^*_{T^{*}})  > \min\{\frac{c_{1}c_{2}c_{3}}{2},
c_{2}c_{4}\}
\end{equation*}
where $c_{1}, c_{2}, c_{3}, c_{4}$ are defined in Assumption
\ref{assump4}.  Moreover, the Go-CART estimator in both the
penalized risk minimization and held-out risk minimization form is
tree partition consistent.
\end{theorem}

This result shows that, with high probability, we obtain a finer
partition than $T^{*}$; the assumptions do not, however, control the
size of the resulting partition.  The proof of this result appears
in the appendix.

\section{Experimental Results}
\label{sec::experiment}

We evaluate the performance of the greedy Go-CART learning algorithm
in Section \ref{sec:greedyalgo} on both synthetic
datasets and a meteorological dataset. In each experiment,  we
set the dyadic integer to $N=2^{10}$ to ensure that we can
obtain fine-tuned partitions of the input space $\mathcal{X}$.
Furthermore, we always ensure that the region (hyperrectangle)
represented by each leaf node contains at least 10 data points
to guarantee reasonable estimates of the sample means and sparse
inverse covariance matrices.

\subsection{Synthetic Data}
\label{sec:synexp}

\begin{figure}[!t]
\centering
\begin{tabular}{cccc}
\includegraphics[width=.25\textwidth]{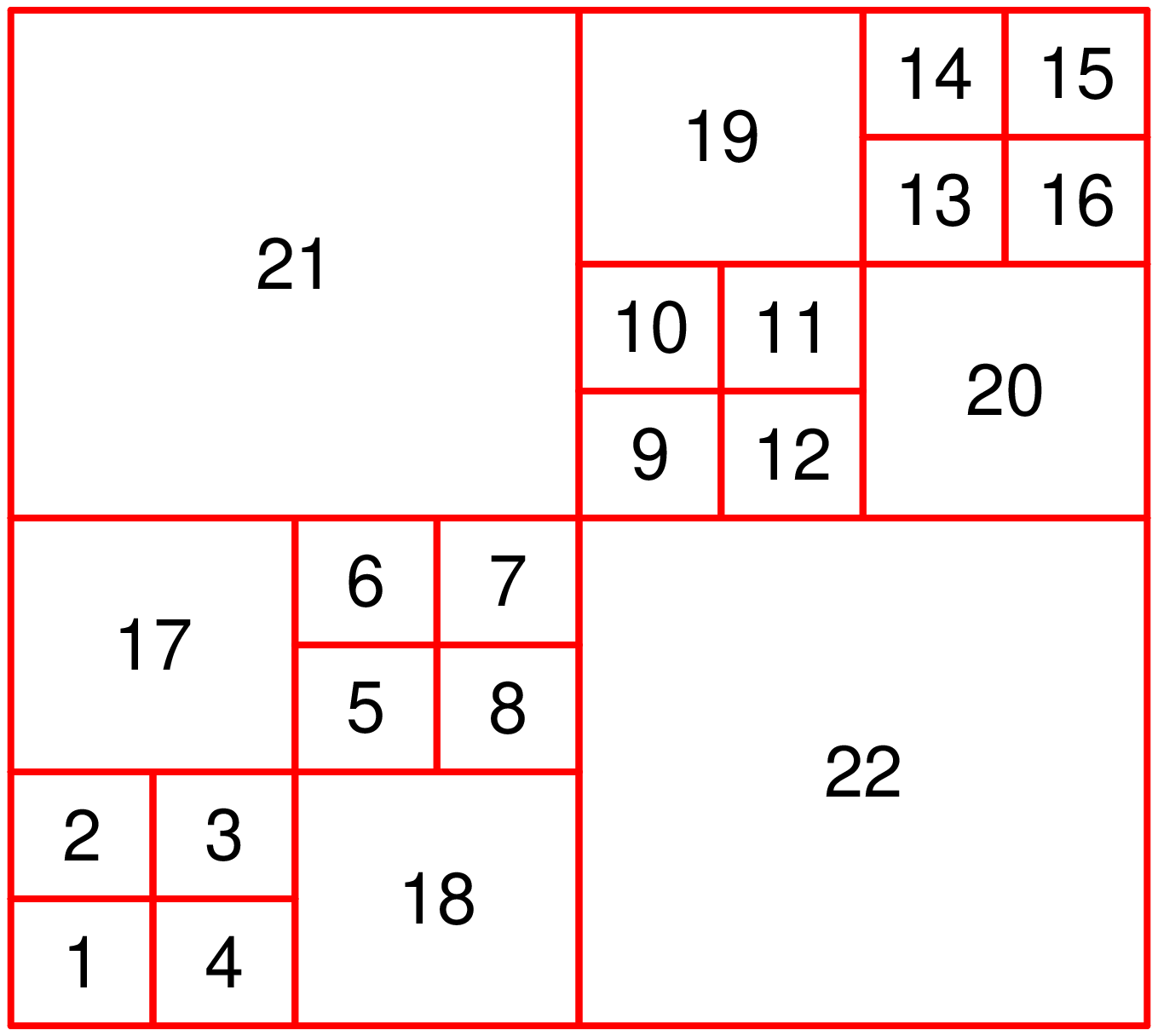} &
\hspace{-0.3in}   
\includegraphics[angle=-90,width=.25\textwidth,origin=0]{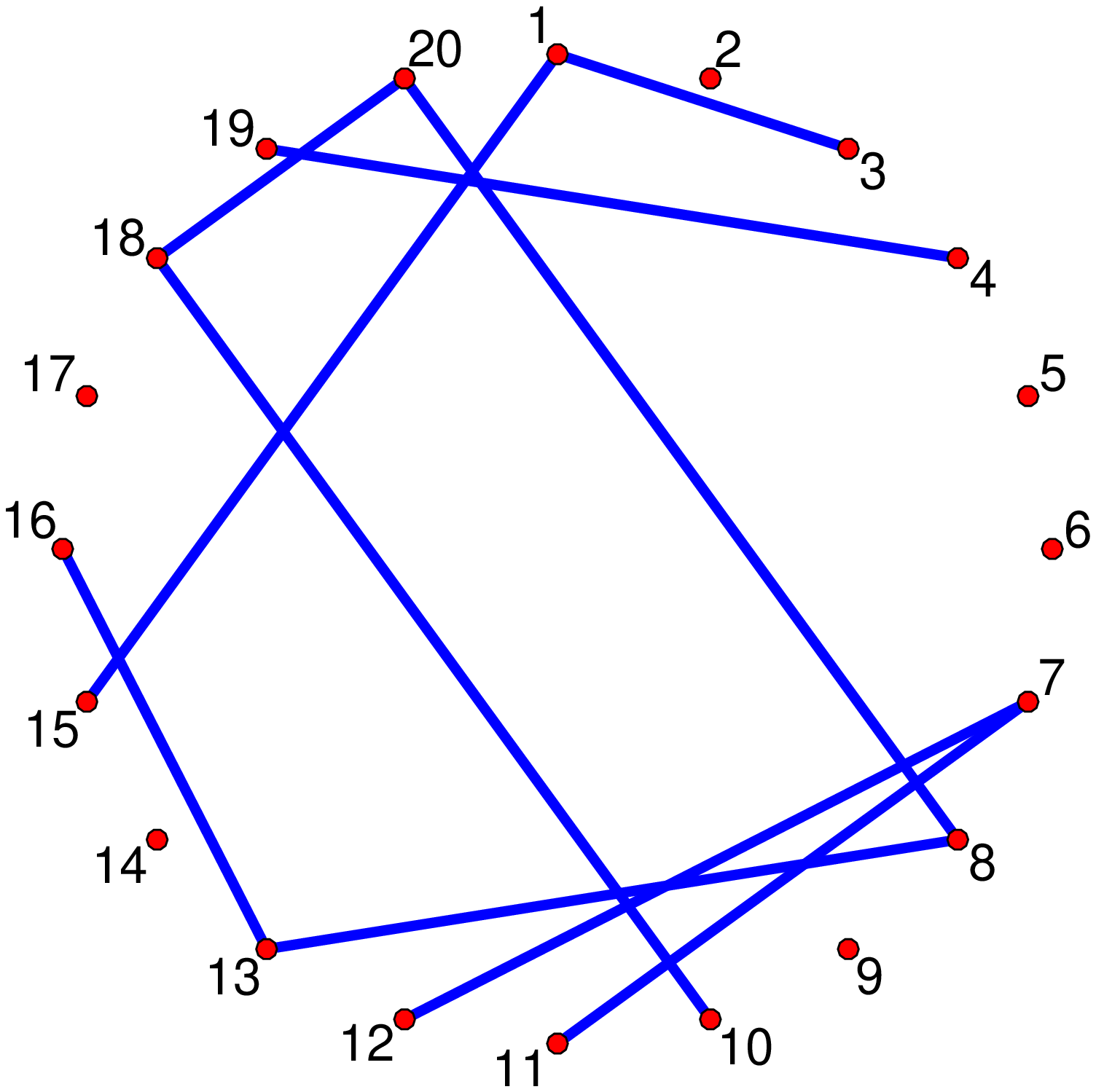} & 
\hspace{-0.3in}
\includegraphics[angle=-90, width=.25\textwidth,origin=0]{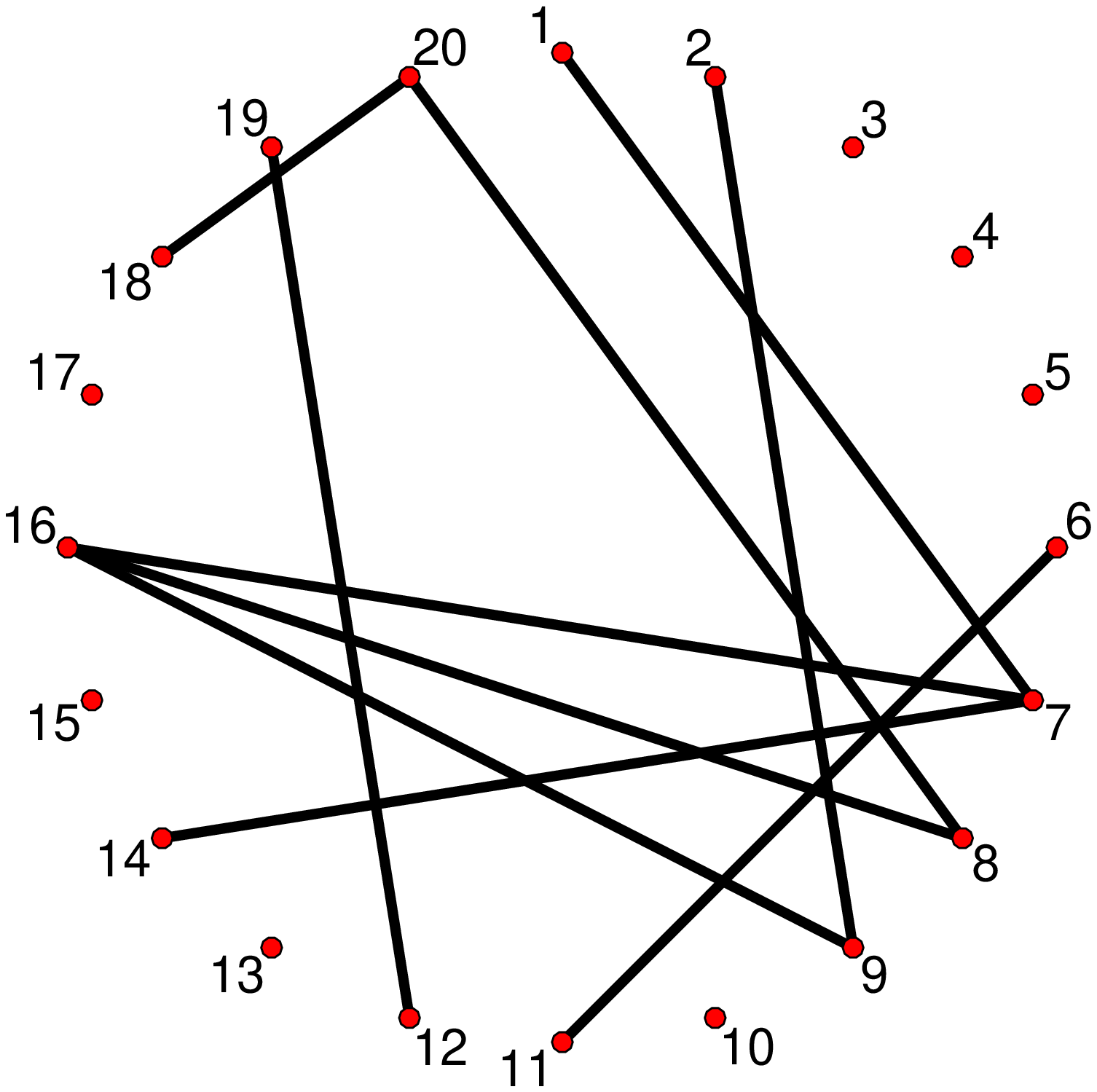} &
\hspace{-0.3in}  
\includegraphics[angle=-90, width=.25\textwidth, origin=0]{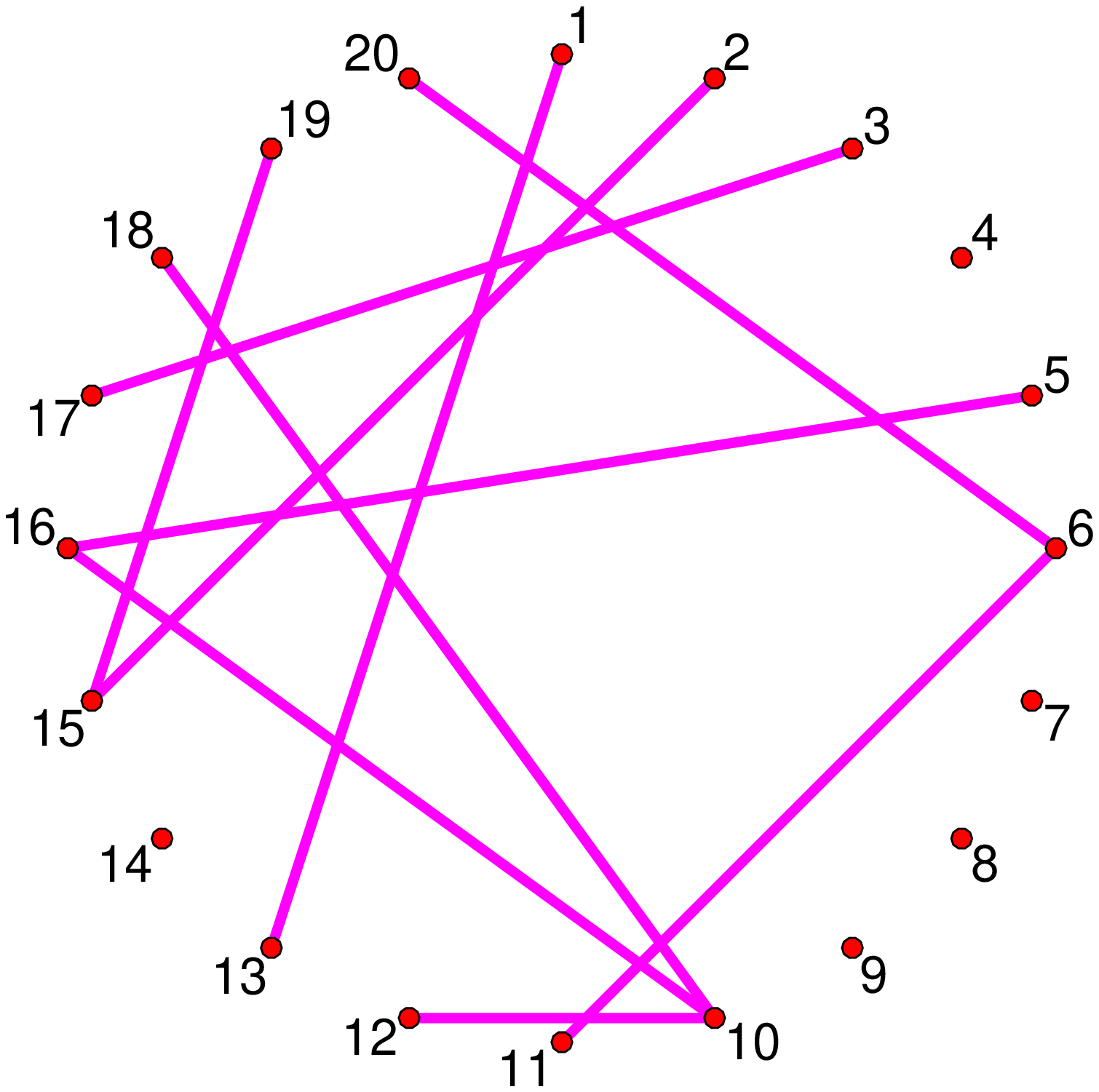}
\vspace{-0.3cm}  \\
\vspace{-0.3cm}   (a) & 
\hskip-10pt(b) & 
\hskip-10pt(c) & 
\hskip-10pt(d)
\end{tabular}
\vskip15pt
\caption{\small (a) The 22 subregions defined on $[0,1]^2$. The
  horizontal axis corresponds to the first dimension denoted as $X_1$
  while the vertical axis corresponds to the second dimension denoted
  as $X_2$. The bottom left point corresponds to $[0,0]$ and the upper
  right point corresponds to $[1,1]$.  (b) The true graph for
  subregion 4. (c) The true graph for subregion 17. (d) The
  true graph for subregion 22.}
   \label{fig:synpart}
\end{figure}

We generate $n$ data points $x_1, \ldots,x_n \in \mathbb{R}^{d}$
with $n=10,000$ and $d=10$ uniformly distributed on the unit
hypercube $[0,1]^{d}$. We split the square $[0,1]^2$ defined by the
first two dimensions into 22 subregions, as
shown in Figure \ref{fig:synpart}(a). For the $t$-th subregion
where $1 \leq t \leq 22$, we generate an Erd\"os-R\'enyi random
graph $G^t=(V^t,E^t)$ with $p=20$ vertices and
$|E|=10$ edges, with maximum node degree four. As an
illustration,  the random graphs for subregion four (the smallest
region), 17 (middle region) and 22 (large region) are presented in
Figures \ref{fig:synpart}(b), (c) and (d), respectively. For each
graph $G^t$, we generate an inverse covariance matrix $\Omega^t$
according to:
  \begin{equation*}
    \Omega^{t}_{i,j} = \begin{cases}\
                            1 &  \mathrm{if}~i=j, \\
                            0.245 & \mathrm{if}~(i,j)\in E^t, \\
                            0 & \mathrm{otherwise,}
             \end{cases}
  \end{equation*}
where $0.245$ guarantees positive-definiteness of $\Omega^t$
when the maximum node degree is four.

To each data point  $x_i$ in the $t$-th subregion we associate 
a 20-dimensional response vector $y_i$ generated from a
multivariate Gaussian distribution $N_{20}\left(0, \bigl(
\Omega^{t}\right)^{-1}\bigr)$.  We also create an equally-sized
held-out dataset in the same manner based on
$\{\Omega^t\}_{t=1}^{22}$.

\begin{figure}[t]
\centering
  \begin{small}
  \includegraphics[scale=0.85]{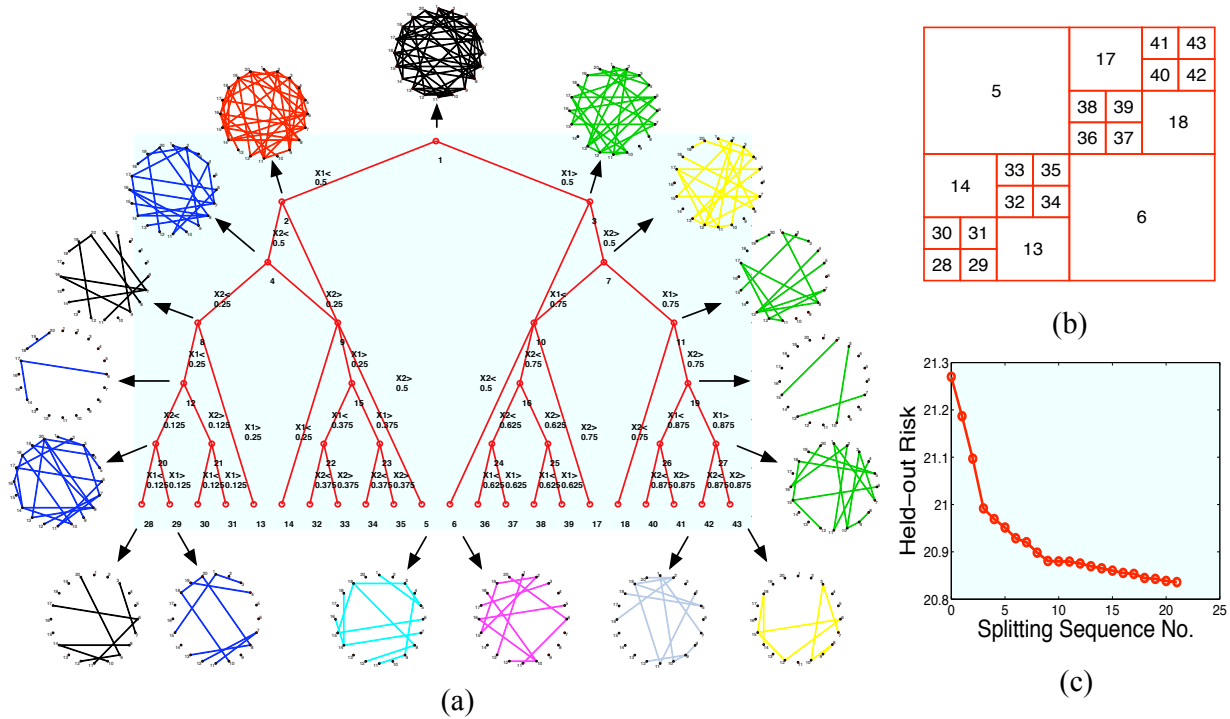}
  \end{small}
\vskip-10pt
  \caption{ (a) The estimated dyadic tree structure; (b) the induced
    partition on $[0,1]^2$ and the number labeled on each subregion
    corresponds to each leaf node ID of the tree in (a); (c) the
    held-out negative log-likelihood risk for each split. The order of
    the splits corresponds the ID of the tree node (from small to
    large)}
   \label{fig:syntree}
\end{figure}

We apply Algorithm \ref{algo:treegrow} to this synthetic dataset.
The estimated dyadic tree structure and its induced partitions are
presented in Figure \ref{fig:syntree}. Estimated graphs for some
nodes are also illustrated.  Note that the label for each subregion
in subplot (c) is the leaf node ID of the tree in subplot (a). We
conduct 100 Monte-Carlo  simulations and find that in 82 out of
100 runs our algorithm perfectly recovers the ground truth partition
of the $X_{1}$-$X_{2}$ plane, and never wrongly splits on any of
the irrelevant dimensions, ranging from $X_{3}$ to $X_{10}$. Moreover, the estimated
graphs have interesting patterns.  Even though the graphs within
each subregion are sparse, the estimated graph obtained by pooling all the
data together is highly dense. As the algorithm progresses,
the estimated graphs become more sparse. However, for the
immediate parent nodes of the true subregions, the graphs become
denser again. 

Out of the 82 simulations where we correctly identify
the tree structure,  we list the graph estimation performance for
subregions 1, 4, 17, 18, 21, 22 in terms of precision, recall, and
$F_1$-score.  Let $\hat{E}$ be the estimated edge set while $E$ be the
true edge set. These criteria are defined as:
\begin{eqnarray}
\text{precision} = \frac{|\hat{E} \cap
E|}{|\hat{E}|},~~\text{recall} = \frac{|\hat{E} \cap
E|}{|E|},~~F_1\text{-score} = 2\cdot\frac{\text{precision}\cdot
\text{recall}}{\text{precision}+ \text{recall}}.
\end{eqnarray}

We see that for a larger subregion, it is easier to obtain
better recovery performance, while good recovery for a very small
region is more challenging.  Of course, in the smaller
regions there is less data. In Figure \ref{fig:synpart}(a), there are only $10000/64
\thickapprox 156$ data points that appear in 
subregion 1 (the smallest one).  In
contrast, approximately $10000/16= 625 $ data
points fall inside subregion 18, so that
the graph corresponding to this region can be better estimated.

We also plot the held-out risk in the subplot (c). As can be seen,
the first few splits lead to the most significant decrease in
the held-out risk.  

Further simulations 
where the ground truth covariance matrix is a continuous
function of~$x$ are presented in the appendix.

\newcolumntype {Q}{>{$\displaystyle}l<{$}}
\newcolumntype {A}{>{$}c <{$}}
\begin{table}[!t]
 \caption{\small Graph estimation performance over different subregions}
{\sf \footnotesize
\begin{tabular}{QAAAAAA}
\toprule %
\multicolumn{6}{c}{ Mean values over 100 runs (Standard deviation)}  \\
\cmidrule(r){2-7}
   \mathbf{subregion}       & \text{region 1}  & \text{region 4} & \text{region 17} & \text{region 18}  &  \text{region 21} &  \text{region 22} \\
\midrule \mathrm{Precision} & 0.8327 \ (0.15) &   0.8429 \ (0.15) & 0.9821 \ (0.05) &   0.9853 \ (0.04) & 0.9906\ (0.04) &0.9899\ (0.05) \\
\mathrm{Recall} & 0.7890 \ (0.16) & 0.7990 \ (0.18) & 1.0000 \ (0.00) &  1.0000 \ (0.00) & 1.0000\ (0.00) & 1.0000 \ (0.00)  \\
F_1-\mathrm{score} & 0.7880 \ (0.11) & 0.7923 \ (0.12) & 0.9904 \
(0.03) &  0.9921 \ (0.02) & 0.9949\ (0.02) & 0.9913\ (0.02)
\\\bottomrule
\end{tabular}}
  \label{tab:seq}
\end{table}

%

%

\subsection{Climate Data Analysis}
\label{sec:Climate}

In this section, we use graph-valued regression to analyze a meteorology dataset
\citep{Lozano:09} that contains monthly data of 18 different
meteorological  factors from 1990 to 2002. We use the data from 1990
to 1995 as the training data and the data from 1996 to 2002 as the
held-out validation data. The observations span 125 locations in the
US on an equally spaced grid between latitude 30.475
and 47.975 and longitude -119.75 to -82.25. The 18
meteorological factors measured for each month include levels of
\textsf{\small CO$_2$}, \textsf{\small CH$_4$}, \textsf{\small
H$_2$}, \textsf{\small CO}, average temperature (\textsf{\small
TMP}) and diurnal temperature range (\textsf{\small DTR}), minimum
temperate (\textsf{\small TMN}), maximum temperature (\textsf{\small
TMX}), precipitation (\textsf{\small PRE}), vapor (\textsf{\small
VAP}), cloud cover (\textsf{\small CLD}), wet days (\textsf{\small
WET}), frost days (\textsf{\small FRS}), global solar radiation
(\textsf{\small GLO}), direct solar radiation (\textsf{\small DIR}),
extraterrestrial radiation (\textsf{\small ETR}), extraterrestrial
normal radiation (\textsf{\small ETRN}) and UV aerosol index
(\textsf{\small UV}). For further detail, see  \cite{Lozano:09}.

\begin{figure}[t]
\centering
  \includegraphics[scale=0.85 , angle=0]{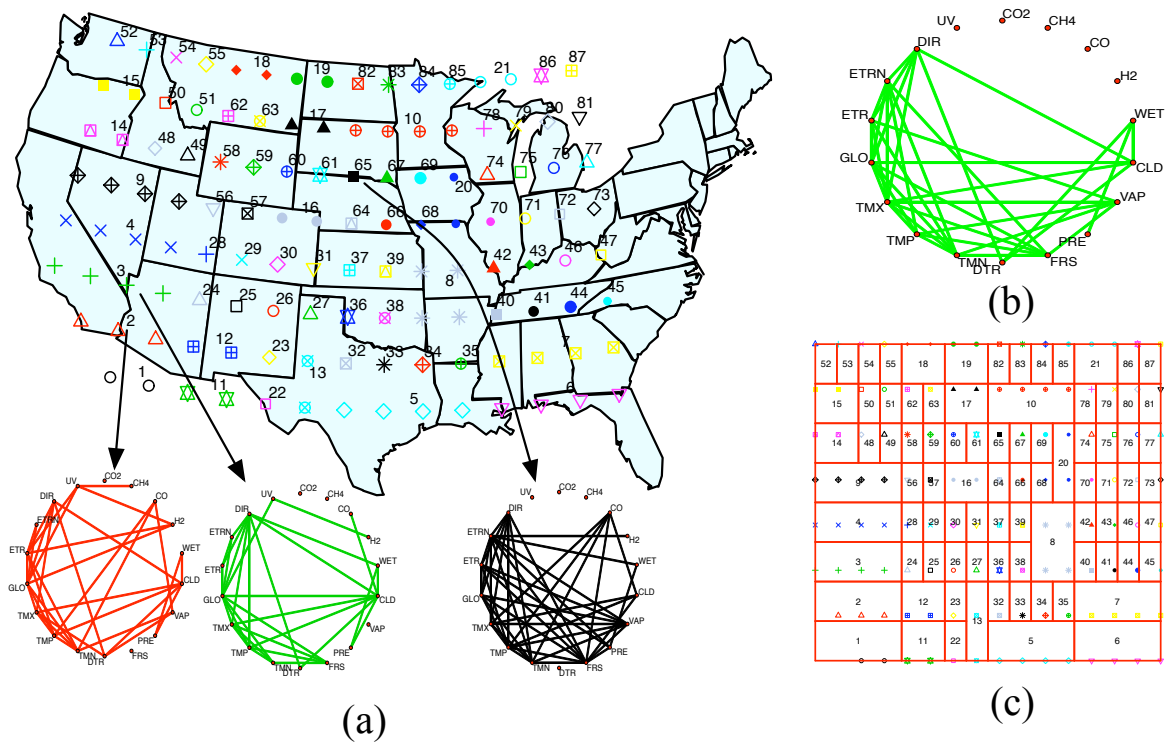}
\vskip-10pt
\caption{\small Analysis of the climate data.  (a) Estimated
  partitions for 125 locations projected to the US map, with the
  estimated graphs for subregions 2, 3, and 65; (b) estimated graph
  with data pooled from all 125 locations; (c) the re-scaled
  partition pattern induced by the dyadic tree structure.}
   \label{fig:climate_graph}
\end{figure}

\begin{figure}[t]
\centering
  \includegraphics[scale=0.55   , angle=0]{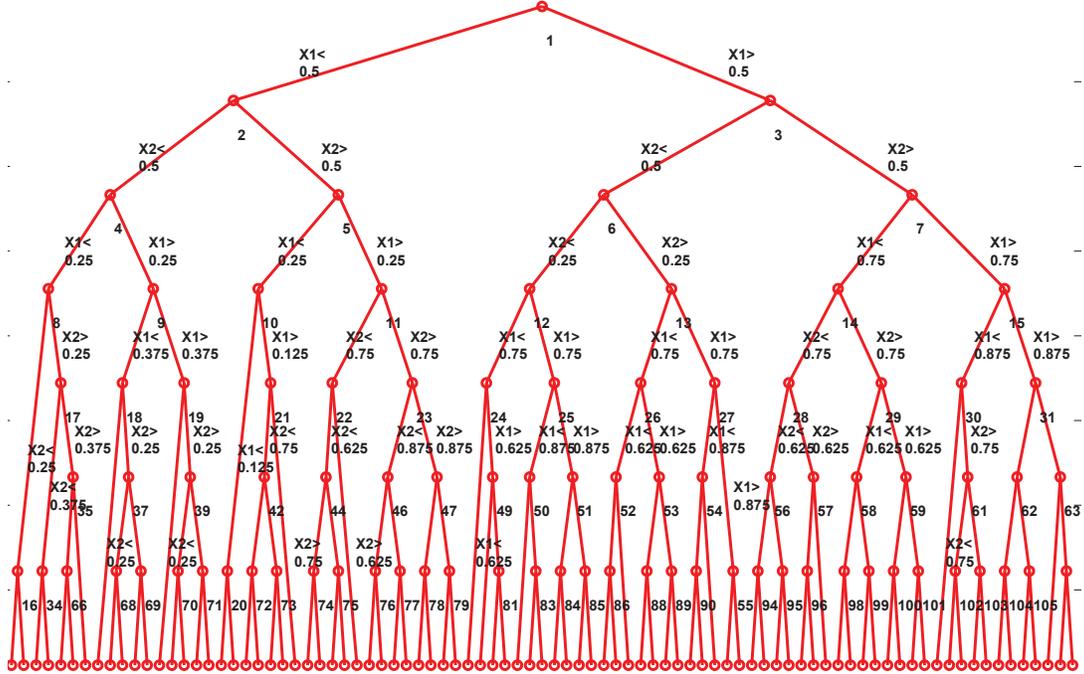}
   \caption{\small  The estimated dyadic tree structure on the climate data.}
   \label{fig:climate_part}
\end{figure}

As a baseline, we estimate a sparse graph on the data from all 125
locations, using the glasso algorithm; the estimated graph
is shown in Figure \ref{fig:climate_graph} (b). It is seen that
there is no edge connecting to any of the greenhouse gas factors
\textsf{\small CO$_2$}, \textsf{\small
  CH$_4$}, \textsf{\small H$_2$} or \textsf{\small CO}.  This
apparently contradicts basic domain knowledge that these four
factors should correlate with the solar radiation factors (including
\textsf{\small GLO}, \textsf{\small DIR}, \textsf{\small
  ETR}, \textsf{\small ETRN}, and \textsf{\small UV}), according to
the 2007 report of the Intergovermental Panel on Climate Change
\cite{IPCC}.
The reason for the missing edges in
the pooled data may be that positive correlations at one location
are canceled by negative correlations at other locations.

Treating the longitude and latitude of each site as two-dimensional
covariate $X$, and the meteorology data of the $p=18$ factors as the
response $Y$, we estimate a dyadic tree structure using the greedy
algorithm.  The result is a partition with 87 subregions, shown in
Figure~\ref{fig:climate_graph}, with the corresponding dyadic partition tree is shown in Figure \ref{fig:climate_part}.
The graphs for subregion 2
(corresponding to the strip of land from Los Angeles, California to
Phoenix, Arizona) and subregion 3 (Bakersfield, California to
Flagstaff, Arizona) are shown in subplot (a) of Figure
\ref{fig:climate_graph}.  The graphs for these two adjacent
subregions are quite similar, suggesting spatial smoothness of the
learned graphs.  Moreover, for both graphs, \textsf{\small CO} is
connected to solar radiation factors in either a direct or indirect
way, and \textsf{\small H$_2$} is connected to $\textsf{\small UV}$,
which is accordance with Chapter 7 of the IPCC report \cite{IPCC}.  In
contrast, for subregion 65, which corresponds to the
border of South Dakota and Nebraska; here the graph is quite
different.   In general, it is found that the graphs corresponding
to the locations along the coasts are sparser than those
corresponding to more central locations in the mainland.  

Such observations,
which require validation and interpretation by domain experts, are
examples of the capability of graph-valued regression to
provide a useful tool for high dimensional data analysis.

\section{Conclusions}
\label{sec::conclusion}

In this paper, we present Go-CART, a partition-based estimator of the family of
undirected graphs associated with a high dimensional conditional
distribution.  Dyadic partitioning estimators, either using penalized
empirical risk minimization or data splitting, are attractive due to
their simplicity and theoretical guarantees.  We derive
finite sample oracle inequalities on excess risk, together with a tree
partition consistency result.  Our theory allows the scale of the
graphs to increase with the sample size, which is relevant since the
methods are targeted at high dimensional data analysis applications.
Greedy partitioning estimators are proposed that are computationally
attractive, combining classical greedy algorithms for decision
trees with recent advances in $\ell_1$-regularization techniques for
graph selection.  The practical potential of Go-CART is indicated by
experiments on a meteorology dataset.  A theoretical analysis of 
greedy Go-CART is one of several interesting directions for future
work.

\appendix

\section{Proofs of Technical Results}
\label{append::theory}

\subsection{Proof of Theorem~\ref{thm.oracle}}

For any $T\in \mathcal{T}_{N}$, we denote
\begin{eqnarray}
S_{j,n} = \frac{1}{n}\sum_{i=1}^{n}(y_{i}-\mu_{\mathcal{X}_{j}}) (y_{i}-\mu_{\mathcal{X}_{j}})^{T} \cdot I(x_{i}\in\mathcal{X}_{j}) \\
\bar{S}_{j} = \mathbb{E}(Y-\mu_{\mathcal{X}_{j}})
(Y-\mu_{\mathcal{X}_{j}})^{T} \cdot I(X\in\mathcal{X}_{j}).
\end{eqnarray}
We then have
\begin{eqnarray}
\lefteqn{\left|R(T, \mu_{T}, \Omega_{T}) - \hat{R}(T, \mu_{T}, \Omega_{T}) \right|} \nonumber \\
& \leq & \biggl|\sum_{j=1}^{m} \mathrm{tr}\left[ \Omega_{\mathcal{X}_{j}}\left( S_{j,n} - \bar{S}_{j}\right) \right]\biggr|  + \biggl|\sum_{j=1}^{m}\log |\Omega_{\mathcal{X}_{j}}|\cdot\Bigl[ \frac{1}{n}\sum_{i=1}^{n}I(x_{i}\in\mathcal{X}_{j}) - \mathbb{E}I(X \in \mathcal{X}_{j}) \Bigr]\biggr| \\
& \leq & \underbrace{\sum_{j=1}^{m} \|
\Omega_{\mathcal{X}_{j}}\|_{1} \cdot \left\|  S_{j,n} - \bar{S}_{j}
\right\|_{\infty}}_{A_{1}}  + \underbrace{\sum_{j=1}^{m}\Bigl|\log
|\Omega_{\mathcal{X}_{j}}|\Bigr|\cdot
\biggl|\frac{1}{n}\sum_{i=1}^{n}I(x_{i}\in\mathcal{X}_{j}) -
\mathbb{E}I(X \in \mathcal{X}_{j})\biggr|}_{A_{2}}.
\end{eqnarray}
We now analyze the terms $A_{1}$ and $A_{2}$ separately.

For $A_{2}$, using the Hoeffding's inequality, for $\epsilon >0$, we
get
\begin{equation}
\mathbb{P}\left(\biggl|\frac{1}{n}\sum_{i=1}^{n}I(x_{i}\in\mathcal{X}_{j})
- \mathbb{E}I(X \in \mathcal{X}_{j})\biggr| > \epsilon\right) \leq
2\exp\left(-2n\epsilon^{2} \right),
\end{equation}
which implies that,
\begin{eqnarray}
\mathbb{P}\Biggl(\sup_{T \in \mathcal{T}_{N}}
\biggl|\frac{1}{n}\sum_{i=1}^{n}I(x_{i}\in\mathcal{X}_{j}) -
\mathbb{E}I(X \in \mathcal{X}_{j})\biggr| /\epsilon_{T} > 1\Biggr)
\leq 2\sum_{T \in
\mathcal{T}_{N}}\exp\left(-2n\epsilon^{2}_{T}\right),
\end{eqnarray}
where $\epsilon_{T}$ means $\epsilon$ is a function of $T$. For any
$\delta\in(0,1)$, we have, with probability at least $1 - \delta/4$,
\begin{eqnarray}
\forall T \in \mathcal{T}_{N},
~\biggl|\frac{1}{n}\sum_{i=1}^{n}I(x_{i}\in\mathcal{X}_{j}) -
\mathbb{E}I(X \in \mathcal{X}_{j})\biggr| \leq
\sqrt{\frac{[[T]]\log 2 + \log(8/\delta)}{2n}}
\end{eqnarray}
where $[[T]]>0$ is the prefix code of $T$ given in
\eqref{eq.prefixcode}.

From Assumption \ref{assump1}, since $\Omega_{\mathcal{X}_{j}} \in
\Lambda_{j}$, we have that
\begin{eqnarray}
\max_{1 \leq j \leq m_{T}}\log \left|\Omega_{\mathcal{X}_{j}}
\right| \leq     L_{n}
\end{eqnarray}
Therefore, with probability at least $1 -\delta/4$,
\begin{equation}
A_{2} \leq L_{n} m_{T}\sqrt{\frac{[[T]]\log 2 +
\log(8/\delta)}{2n}}.
\end{equation}
Next, we  analyze the term $A_{1}$.  Since
\begin{equation}
\max_{1 \leq j \leq m_{T}} \| \Omega_{\mathcal{X}_{j}}\|_{1}  \leq
L_{n}. \label{eq::L1bound}
\end{equation}
we only need to bound the term $ \left\|  S_{j,n} - \bar{S}_{j}
\right\|_{\infty}$. By Assumption \ref{assump2} and the union bound,
we have, for any $\epsilon > 0$,
\begin{eqnarray}
\lefteqn{\mathbb{P}\left( \left\|  S_{j,n} - \bar{S}_{j}  \right\|_{\infty} > \epsilon \right)}  \nonumber \\
& \leq & \mathbb{P}\biggl(\Bigl\|\frac{1}{n}\sum_{i=1}^{n} y_{i}y^{T}_{i}I(x_{i} \in \mathcal{X}_{j}) - \mathbb{E}\bigl[YY^{T}I(X\in\mathcal{X}_{j})\bigr]\Bigr\|_{\infty} >\frac{\epsilon}{4}  \biggr)  \label{eq::2ndmomment}\\
&  & + \mathbb{P}\biggl(\Bigl\|\frac{1}{n}\sum_{i=1}^{n} y_{i}\mu^{T}_{\mathcal{X}_{j}}I(x_{i} \in \mathcal{X}_{j}) - \mathbb{E}\bigl[Y\mu^{T}_{\mathcal{X}_{j}}I(X\in\mathcal{X}_{j})\bigr] \Bigr\|_{\infty} >\frac{\epsilon}{4}  \biggr) \label{eq::firstmomdent1} \\
&  & + \mathbb{P}\biggl(\Bigl\|\frac{1}{n}\sum_{i=1}^{n} \mu_{\mathcal{X}_{j}}y^{T}_{i}I(x_{i} \in \mathcal{X}_{j}) - \mathbb{E}\bigl[\mu_{\mathcal{X}_{j}}Y^{T}I(X\in\mathcal{X}_{j})\bigr] \Bigr\|_{\infty} >\frac{\epsilon}{4}  \biggr) \label{eq::firstmomdent2} \\
& &  + \mathbb{P}\biggl(\Bigl\|\frac{1}{n}\sum_{i=1}^{n}
\mu_{\mathcal{X}_{j}} \mu^{T}_{\mathcal{X}_{j}} I(x_{i} \in
\mathcal{X}_{j}) -
\mathbb{E}\bigl[\mu_{\mathcal{X}_{j}}\mu^{T}_{\mathcal{X}_{j}}I(X\in\mathcal{X}_{j})\bigr]
\Bigr\|_{\infty} >\frac{\epsilon}{4}  \biggr). \label{eq::mubound}
\end{eqnarray}
Using the fact that  $\|\mu\|_{\infty} \leq B$ and  the Assumption
\ref{assump2}, we can apply Bernstein's exponential inequality on
\eqref{eq::2ndmomment}, \eqref{eq::firstmomdent1}, and
\eqref{eq::firstmomdent2}.   Also, since the indicator function is
bounded, we can apply Hoeffding's inequality on \eqref{eq::mubound}.
In this way we obtain
\begin{eqnarray}
\lefteqn{\mathbb{P}\left( \left\|  S_{j,n} - \bar{S}_{j}  \right\|_{\infty} > \epsilon \right)} \\
& &\leq 2 p^{2}\exp\left(-\frac{1}{32}\biggl(
\frac{n\epsilon^{2}}{v_{2} + M_{2}\epsilon}\biggr) \right)   +
4p^2\exp\left(-\frac{1}{32B^{2}}\biggl(  \frac{n\epsilon^{2}}{v_{1}
+ M_{1}\epsilon}\biggr) \right) + 2p^{2} \exp\left(-\frac{2
n\epsilon^{2}}{B^{4}} \right). \nonumber
\end{eqnarray}
Therefore, for any $\delta\in(0,1)$, we have, for any $\epsilon
\rightarrow 0$ as $n$ goes to infinity,  with probability at least
$1 - \delta/4$
\begin{eqnarray}
\forall T \in \mathcal{T}_{N},  ~ \left\|  S_{j,n} - \bar{S}_{j}  \right\|_{\infty} &\leq  & (8 \sqrt{v_{2}})\cdot\sqrt{\frac{[[T]]\log 2 +2\log p +\log(24/\delta)}{n}} \\
 & + & (8 B \sqrt{v_{1}})\cdot\sqrt{\frac{[[T]]\log 2 +2\log p +\log(48/\delta)}{n}} \\
 & + & B^{2}\cdot\sqrt{\frac{[[T]]\log 2 +2\log p +\log(24/\delta)}{2n}}
\end{eqnarray}
Combined with \eqref{eq::L1bound}, we get that
\begin{eqnarray}
A_{1} \leq C_{1} L_{n} m_{T} \sqrt{\frac{[[T]]\log 2 +2\log p
+\log(48/\delta)}{n}}.
\end{eqnarray}
where $C_{1} = 8 \sqrt{v_{2}} + 8 B \sqrt{v_{1}}+ B^{2}$.

Since the above analysis holds uniformly over 
$\mathcal{T}_{N}$,  when choosing
\begin{eqnarray}
\mathrm{pen}(T) = (C_{1}+1) L_{n}m_{T} \sqrt{\frac{[[T]]\log 2
+2\log p +\log(48/\delta)}{n}},
\end{eqnarray}
we then get, with probability at least $1 - \delta/2$,
\begin{eqnarray}
\sup_{T\in\mathcal{T}_{N}, \mu_{j}\in M_{j}, \Omega_{j}\in
\Lambda_{j}  }\left|R(T, \mu_{T}, \Omega_{T}) - \hat{R}(T, \mu_{T},
\Omega_{T}) \right| \leq \mathrm{pen}(T)
\label{eq::uniformdeviation}
\end{eqnarray}
for large enough $n$.

Given a DPT $T$, we define
\begin{equation}
\mu^{o}_{T}, \Omega^{o}_{T} = \argmin_{\mu_{T}\in M_{j},\Omega_{T}\in\Lambda_{j}} R(T, \mu_{T}, \Omega_{T}).
\end{equation}
From the uniform deviation inequality in
\eqref{eq::uniformdeviation}, we have, for large enough $n$: for any
$\delta \in (0,1)$, with probability at least $1 - \delta$,
\begin{eqnarray}
R(\hat{T},\hat{\mu}_{\hat{T}}, \hat{\Omega}_{\hat{T}}) & \leq  & \hat{R}(\hat{T},\hat{\mu}_{\hat{T}}, \hat{\Omega}_{\hat{T}})  + \mathrm{pen}(\hat{T}) \\
& = & \inf_{T\in \mathcal{T}_{N}, \mu_{\mathcal{X}_j}\in M_j, \Omega_{\mathcal{X}_j} \in \Lambda_{j}}\biggl\{ \hat{R}(T, \mu_{T}, \Omega_{T}) + \mathrm{pen}(T)\biggr\} \\
& \leq & \inf_{T\in\mathcal{T}_{N}}\left\{ \hat{R}(T, \mu^{0}_{T}, \Omega^{0}_{T}) + \mathrm{pen}(T) \right\} \\
& \leq & \inf_{T\in\mathcal{T}_{N}}\left\{ {R}(T, \mu^{0}_{T}, \Omega^{0}_{T}) + 2\mathrm{pen}(T) \right\} \\
& = &\inf_{T\in\mathcal{T}_{N}}\left\{ \inf_{\mu_{\mathcal{X}_j}\in
M_{j}, \Omega_{\mathcal{X}_j} \in \Lambda_{j}}(R(T, \mu_{T},
\Omega_{T})  + 2\mathrm{pen}(T) \right\}.
\end{eqnarray}
The desired result of the theorem follows by subtracting $R^{*}$
from  both sides.

\subsection{Proof of Theorem~\ref{thm.oracle2}}

From  \eqref{eq::uniformdeviation} we have, for large enough $n$,
on the dataset $\mathcal{D}_{1}$, with probability at least $1 -
{\delta}/{4}$
\begin{eqnarray}
\sup_{T\in\mathcal{T}_{N}, \mu_{j}\in M_{j}, \Omega_{j}\in
\Lambda_{j}  }\left|R(T, \mu_{T}, \Omega_{T}) - \hat{R}(T, \mu_{T},
\Omega_{T}) \right| \leq \phi_{n}(T). \label{eq::uniformdeviation1}
\end{eqnarray}
Following the same line of analysis, we can also get that on the validation
dataset $\mathcal{D}_{2}$, with probability at least
$1-\delta/4$,
\begin{eqnarray}
\sup_{T\in\mathcal{T}_{N}}\left|R(T, \hat{\mu}_{T},
\hat{\Omega}_{T}) - \hat{R}_{\rm out}(T, \hat{\mu}_{T},
\hat{\Omega}_{T}) \right| \leq \phi_{n}(T)
\label{eq::uniformdeviation2}
\end{eqnarray}
for large enough $n$. Here $\hat{\mu}_{T}, \hat{\Omega}_{T}$ are as
defined in \eqref{eq::step1est}.

Given a DPT $T$, we define
\begin{equation}
\mu^{o}_{T}, \Omega^{o}_{T} = \argmin_{\mu_{T}\in M_{j},\Omega_{T}\in\Lambda_{j}} R(T, \mu_{T}, \Omega_{T}).
\end{equation}

Using the fact that
\begin{eqnarray}
\hat{T}= \ds {\rm argmin}_{T\in  \mathcal{T}_{N}}\hat{R}_{\rm
out}(T, \hat{\mu}_{T}, \hat{\Omega}_{T}),
\end{eqnarray}
we have, for large enough $n$ and any $\delta \in (0,1)$, with
probability at least $1 - \delta$,
\begin{eqnarray}
R(\hat{T},\hat{\mu}_{\hat{T}}, \hat{\Omega}_{\hat{T}}) & \leq  & \hat{R}_{\rm out}(\hat{T},\hat{\mu}_{\hat{T}}, \hat{\Omega}_{\hat{T}}) + \phi_{n}(\hat{T}) \\
& = & \inf_{T\in \mathcal{T}_{N}}  \hat{R}_{\rm out}(T, \hat{\mu}_{T}, \hat{\Omega}_{T}) + \phi_{n}(\hat{T})\\
& \leq & \inf_{T\in \mathcal{T}_{N}}\left\{ R(T, \hat{\mu}_{T}, \hat{\Omega}_{T}) + \phi_{n}(T) \right\} +\phi_{n}(\hat{T}) \\
& \leq & \inf_{T\in \mathcal{T}_{N}}\left\{ \hat{R}(T, \hat{\mu}_{T}, \hat{\Omega}_{T}) + \phi_{n}(T) + \phi_{n}(T) \right\} +\phi_{n}(\hat{T}) \\
& \leq & \inf_{T\in \mathcal{T}_{N}}\left\{ \hat{R}(T, \mu^{0}_{T}, \Omega^{0}_{T}) + \phi_{n}(T) + \phi_{n}(T) \right\} +\phi_{n}(\hat{T}) \\
& \leq &\inf_{T \in  \mathcal{T}_{N}}\left\{  3 \phi_{n}(T) +
\inf_{\mu_{\mathcal{X}_{j}}\in M_{j}, \Omega_{\mathcal{X}_{j}} \in
\Lambda_{j}}R(T, \mu_{T}, \Omega_{T}) \right\} + \phi_{n}(\hat{T}).
\nonumber
\end{eqnarray}
The result follows by subtracting $R^{*}$ from both sides.

\subsection{Proof of Theorem~\ref{thm.sparsistent}}

  For any $T\in \mathcal{T}_{N}$, $\Pi(T^{*}) \nsubseteq \Pi(T) $,
  there must exist a subregion $\mathcal{X}'\in \Pi(T)$ such that
  no $\mathcal{A} \in \Pi(T^{*})$ satisfies
  $\mathcal{X}' \subset \mathcal{A}$.  We can thus find a
  minimal class of disjoint subregions $\{\mathcal{X}^{0}_{1},
  \ldots, \mathcal{X}^{0}_{k'}\} \in \Pi(T^{*})$, such that
\begin{eqnarray}
\mathcal{X}' \subset \cup_{i=1}^{k'}\mathcal{X}^{0}_{i},
\end{eqnarray}
where $k' \geq 2$. We define $\mathcal{X}^{*}_{i} = X^{0}_{i}
\cap \mathcal{X}'$ for $i=1,\ldots, k'$.  Then we have
\begin{equation}
\mathcal{X}' = \cup_{i=1}^{k'}\mathcal{X}^{*}_{i}.
\end{equation}

Let $\{ \mu^*_{\mathcal{X}^{*}_j}, \Omega^*_{\mathcal{X}^{*}_j}
\}_{j=1}^{k'}$ be the  true parameters on $
\mathcal{X}^{0}_{1}, \ldots,  \mathcal{X}^{0}_{k'}$. We
denote by $R(\mathcal{X}', \mu^*_{T^{*}}, \Omega^*_{T^{*}} )$ the
risk of $\mu^*_{T^{*}}$ and $\Omega^*_{T^{*}}$ on the subregion
$\mathcal{X}'$, so that
\begin{eqnarray}
R(\mathcal{X}', \mu^*_{T^{*}}, \Omega^*_{T^{*}}) &=& \sum_{j=1}^{k'}\mathbb{E}\biggl[\left( \mathrm{tr}\left[ \Omega^*_{\mathcal{X}^{*}_j}\bigl( (Y-\mu^{*}_{\mathcal{X}^{*}_j}) (Y-\mu^{*}_{\mathcal{X}^{*}_j})^{T} \bigr) \right]  - \log |\Omega^*_{\mathcal{X}^{*}_j}|\right)\cdot I(X \in \mathcal{X}^{*}_{j})\biggr] \nonumber\\
&= & p\mathbb{P}\left(X \in \mathcal{X}' \right) -
\sum_{j=1}^{k'}\mathbb{P}\left(X \in \mathcal{X}^*_j\right)\log
|\Omega^*_{\mathcal{X}^{*}_j}|.
\end{eqnarray}
Since the DPT $T$ does not further partition $\mathcal{X}'$,   we
have, for any $\mu_{T}, \Omega_{T}\in\mathcal{M}_{T}$
\begin{eqnarray}
\lefteqn{R(\mathcal{X}', \mu_{T}, \Omega_{T}) } \nonumber \\
&=& \sum_{j=1}^{k'}\mathbb{E}\biggl[\Bigl( \mathrm{tr}\left[ \Omega_{T}\left( (Y-\mu_{T}) (Y-\mu_{T})^{T} \right) \right]  - \log |\Omega_{T}|\Bigr)\cdot I(X \in \mathcal{X}^{*}_{j}) \biggr]\nonumber \\
& =& \sum_{j=1}^{k'}\mathbb{E}\biggl[\Bigl( \mathrm{tr}\left[
\Omega_{T}\left( (Y-\mu_{T}) (Y-\mu_{T})^{T} \right) \right]
\Bigr)\cdot I(X \in \mathcal{X}^{*}_{j}) \biggr] - \mathbb{P}(X \in
\mathcal{X}')\log|\Omega_{T}|. \nonumber
\end{eqnarray}
Using the decomposition
\begin{eqnarray}
\lefteqn{ (Y-\mu_{T}) (Y-\mu_{T})^{T}  =   (Y-\mu^*_{\mathcal{X}^{*}_j}) (Y-\mu^*_{\mathcal{X}^{*}_j})^{T}  + (Y-\mu^*_{\mathcal{X}^{*}_j})(\mu^*_{\mathcal{X}^{*}_j} - \mu_{T})^{T}  } \nonumber \\
& & ~~~~~~~~~~~~~+ (\mu^*_{\mathcal{X}^{*}_j} -
\mu_{T})(Y-\mu^*_{\mathcal{X}^{*}_j})^{T}
+(\mu^*_{\mathcal{X}^{*}_j} - \mu_{T})(\mu^*_{\mathcal{X}^{*}_j} -
\mu_{T})^{T}
\end{eqnarray}
we obtain
\begin{eqnarray}
\lefteqn{\sum_{j=1}^{k'}\mathbb{E}\biggl[\Bigl( \mathrm{tr}\left[ \Omega_{T}\left( (Y-\mu_{T}) (Y-\mu_{T})^{T} \right) \right] \Bigr)\cdot I(X \in \mathcal{X}^{*}_{j}) \biggr] } \nonumber \\
& = & \sum_{j=1}^{k'}\mathbb{P}\left(X \in
\mathcal{X}^*_j\right)\left[\mathrm{tr}(\Omega_{T}
(\Omega^{*}_{j})^{-1}) +
\mathrm{tr}(\Omega_{T}(\mu^*_{\mathcal{X}^{*}_j} -
\mu_{T})(\mu^*_{\mathcal{X}^{*}_j} - \mu_{T})^{T}) \right].
\end{eqnarray}
Using the bound
\begin{eqnarray}
R(\mathcal{X}', \mu_{T}, \Omega_{T} ) \geq \max\{R(\mathcal{X}',
\mu^*_{T^{*}}, \Omega_{T} ), R(\mathcal{X}', \mu_{T},
\Omega^*_{T^{*}} ) \},
\end{eqnarray}
we proceed by cases.

\noindent
{\bf Case 1}: The $\mu$'s are different.
We know that
\begin{eqnarray}
\lefteqn{\inf_{\mu_{T}, \Omega_{T}\in\mathcal{M}_{T}}R(\mathcal{X}',
  \mu_{T}, \Omega_{T})  - R(\mathcal{X}', \mu^*_{T^{*}},
  \Omega^*_{T^{*}}) } \qquad \ \\
 &  \geq&   \inf_{\mu_{T}} R(\mathcal{X}', \mu_{T}, \Omega^*_{T^{*}})  - R(\mathcal{X}', \mu^*_{T^{*}}, \Omega^*_{T^{*}}) \nonumber \\
&   =&\inf_{\mu_{T}}\sum_{j=1}^{k'}\mathbb{P}\left(X \in \mathcal{X}^*_j\right)(\mu^*_{\mathcal{X}^{*}_j} - \mu_{T})^{T}\Omega^*_{\mathcal{X}^{*}_j}(\mu^*_{\mathcal{X}^{*}_j} - \mu_{T}) \nonumber \\
&  \geq &c_{1}c_{2}
\inf_{\mu_{T}}\sum_{j=1}^{k'}\|\mu^*_{\mathcal{X}^{*}_j} - \mu_{T}
\|^{2}_{2}  \nonumber
\end{eqnarray}
where
 the last inequality follows from that fact that $\rho_{\rm \min}(\Omega^*_{\mathcal{X}^{*}_j}) \geq c_{1} , \mathbb{P}\left(X \in \mathcal{X}^*_j\right) \geq c_{2}$. It's easy to see that a lower bound of the last term is achieved at $\bar{\mu}_{T}$,
\begin{equation}
\bar{\mu}_{T} =
\frac{1}{k'}\sum_{j=1}^{k'}\mu^*_{\mathcal{X}^{*}_j}.
\end{equation}
Furthermore, for any  two DPTs  $T$ and $T'$,  if $\Pi(T) \subset
\Pi(T')$ it's clear that
\begin{eqnarray}
\inf_{\mu_{T}, \Omega_{T}\in\mathcal{M}_{T}}R(T, \mu_{T},
\Omega_{T})  \geq \inf_{\mu_{T'},
\Omega_{T'}\in\mathcal{M}_{T'}}R(T', \mu_{T'}, \Omega_{T'}).
\end{eqnarray}
Therefore,  in the sequel, without loss of generality we only need
to consider the case $k'=2$.

The result in this case then follows from the fact that
\begin{eqnarray}
\sum_{j=1}^{2}\|\mu^*_{\mathcal{X}^{*}_j} - \bar{\mu}_{T} \|^{2}_{2}
= \frac{1}{2}\|\mu_{\mathcal{X}^{*}_{1}} - \mu_{\mathcal{X}^{*}_{2}}
\|^{2}_{2} \geq \frac{c_{3}}{2}.
\end{eqnarray}

\noindent
{\bf Case 2}: The $\Omega$'s are different.
In this case, we have
\begin{eqnarray}
\lefteqn{  \inf_{\mu_{T}, \Omega_{T}\in\mathcal{M}_{T}}R(\mathcal{X}', \mu_{T}, \Omega_{T})  - R(\mathcal{X}', \mu^*_{T^{*}}, \Omega^*_{T^{*}})  \geq  \inf_{\Omega_{T}} R(\mathcal{X}', \mu^*_{T^{*}}, \Omega_{T})  - R(\mathcal{X}', \mu^*_{T^{*}}, \Omega^*_{T^{*}})  } \nonumber \\
& = &\inf_{\Omega_{T}}\sum_{j=1}^{k'}\mathbb{P}\left(X \in \mathcal{X}^*_j\right) \left( \mathrm{tr}\left[\Omega^{-1}_{\mathcal{X}^{*}_j} (\Omega_{T} - \Omega^*_{\mathcal{X}^{*}_j})\right] - \left(\log|\Omega_{T}|  - \log|\Omega^*_{\mathcal{X}^{*}_j}| \right) \right) \\
& \geq &c_{2}\inf_{\Omega_{T}}\sum_{j=1}^{k'}\left( \mathrm{tr}\left[\Omega^{-1}_{\mathcal{X}^{*}_j} (\Omega_{T} - \Omega^*_{\mathcal{X}^{*}_j})\right] - \left(\log|\Omega_{T}|  - \log|\Omega^*_{\mathcal{X}^{*}_j}| \right) \right) \\
& \geq &c_{2}\inf_{\Sigma_{T}}\sum_{j=1}^{k'}\left( \mathrm{tr}\left[\Sigma^*_{\mathcal{X}^{*}_j} (\Sigma^{-1}_{T} - \Omega^*_{\mathcal{X}^{*}_j})\right]  + \log\frac{|\Sigma_{T}|}{|\Sigma^*_{\mathcal{X}^{*}_j}|} \right) \\
& = & c_{2}\inf_{\Sigma_{T}}\sum_{j=1}^{k'}\left(
\mathrm{tr}\left(\Sigma^*_{\mathcal{X}^{*}_j} \Sigma^{-1}_{T}
\right)  + \log\frac{|\Sigma_{T}|}{|\Sigma^*_{\mathcal{X}^{*}_j}|}
-p \right)
\end{eqnarray}
where $\Sigma_{T} = \Omega^{-1}_{T}$.

As before, we only need to consider the case $k'=2$.  A
lower bound of the last term is achieved at
\begin{eqnarray}
\bar{\Sigma}_{T} = \frac{\Sigma_{\mathcal{X}^{*}_{1}} +
\Sigma_{\mathcal{X}^{*}_{2}} }{2}.
\end{eqnarray}
Plugging in $\bar{\Sigma}_{T}$, we get
\begin{eqnarray}
\lefteqn{\inf_{\Sigma_{T}}\sum_{j=1}^{2}\left( \mathrm{tr}\left(\Sigma^*_{\mathcal{X}^{*}_j} \Sigma^{-1}_{T} \right)  + \log\frac{|\Sigma_{T}|}{|\Sigma^*_{\mathcal{X}^{*}_j}|} -p \right)  \geq  \sum_{j=1}^{2}\left( \mathrm{tr}\left(\Sigma^*_{\mathcal{X}^{*}_j} \bar{\Sigma}^{-1}_{T} \right)  + \log\frac{|\bar{\Sigma}_{T}|}{|\Sigma^*_{\mathcal{X}^{*}_j}|} -p \right) } \nonumber \\
&= &   \mathrm{tr}\left((2\bar{\Sigma}_{T} - \Sigma_{\mathcal{X}^{*}_{2}} ) \bar{\Sigma}^{-1}_{T} \right)  + \log\frac{|\bar{\Sigma}_{T}|}{|\Sigma_{\mathcal{X}^{*}_{1}}|} -p  +  \mathrm{tr}\left(\Sigma_{\mathcal{X}^{*}_{2}} \bar{\Sigma}^{-1}_{T} \right)  + \log\frac{|\bar{\Sigma}_{T}|}{|\Sigma_{\mathcal{X}^{*}_{2}}|} -p  \\
& =&  \log\frac{|\bar{\Sigma}_{T}|}{|\Sigma_{\mathcal{X}^{*}_{1}}|}+\log\frac{|\bar{\Sigma}_{T}|}{|\Sigma_{\mathcal{X}^{*}_{2}}|} \\
& = & 2\log\left|\frac{\Sigma_{\mathcal{X}^{*}_{1}} + \Sigma_{\mathcal{X}^{*}_{2}} }{2}\right| - \log|\Sigma_{\mathcal{X}^{*}_{1}}| - \log|\Sigma_{\mathcal{X}^{*}_{2}}|\\
& \geq & c_{4}
\end{eqnarray}
where the last inequality follows from the given assumption.

Therefore, we have
\begin{equation}
\inf_{\mu_{T}, \Omega_{T}\in\mathcal{M}_{T}}R(\mathcal{X}', \mu_{T},
\Omega_{T})  - R(\mathcal{X}', \mu^*_{T^{*}}, \Omega^*_{T^{*}})
\geq c_{2}c_{4}.
\end{equation}
The theorem is obtained by combining the two cases.

\section{Further Simulations}

To further demonstrate the performance of the method,
this section presents simulations where the true conditional
covariance matrix is continuous in $X$. We compare the graphs
estimated by our method to the single graph obtained by applying the glasso directly to
the entire dataset.

In this subsection, we consider the case where $X$ lies on a one
dimensional chain. More precisely, we generate $n$ equally spaced
points $x_1, \ldots,x_n \in \mathbb{R}$ with $n=10,000$ on $[0,1]$.
We generate an Erd\"os-R\'enyi random graph $G^1=(V^1,E^1)$ with 
$p=20$ vertices, $|E|=10$ edges, and 
maximum node degree four.  Then, we simulate the
output $y_1, \ldots,y_n] \in \mathbb{R}^{p}$ as follows:

\begin{enumerate}
  \item For
  $t=2$ to $T$, we construct the graph $G^t=(V^t, E^t)$ as follows:
  (a) with probability 0.05, remove one edge from $G^{t-1}$ and (b)  with probability $0.05$, add one
  edge to the graph generated in (a).
  We make sure that the total number of edges is between 5 and 15, and
  that the maximum node degree four.
  \item For each graph $G^t$, generate the inverse covariance matrix
  $\Omega^t$:
  \begin{equation*}
    \Omega^{t}(i,j) = \begin{cases}\
                            1 &  \mathrm{if}~i=j, \\
                            0.245 & \mathrm{if}~(i,j)\in E^t, \\
                            0 & \mathrm{otherwise,}
             \end{cases}
  \end{equation*}
  where $0.245$ guarantees  positive-definiteness of $\Omega^t$
  under the degree constraint.
  \item For each $t$, we sample $y_t$ from a multivariate
  Gaussian distribution with mean $\mu=(0, \ldots, 0) \in \mathbb{R}^p$ and covariance  matrix $\Sigma^t= (\Omega^t)^{-1}$.
\end{enumerate}

\begin{figure}[t]
\centering
  \begin{small}
  \begin{tabular}{cc}
  \includegraphics[scale=0.39]{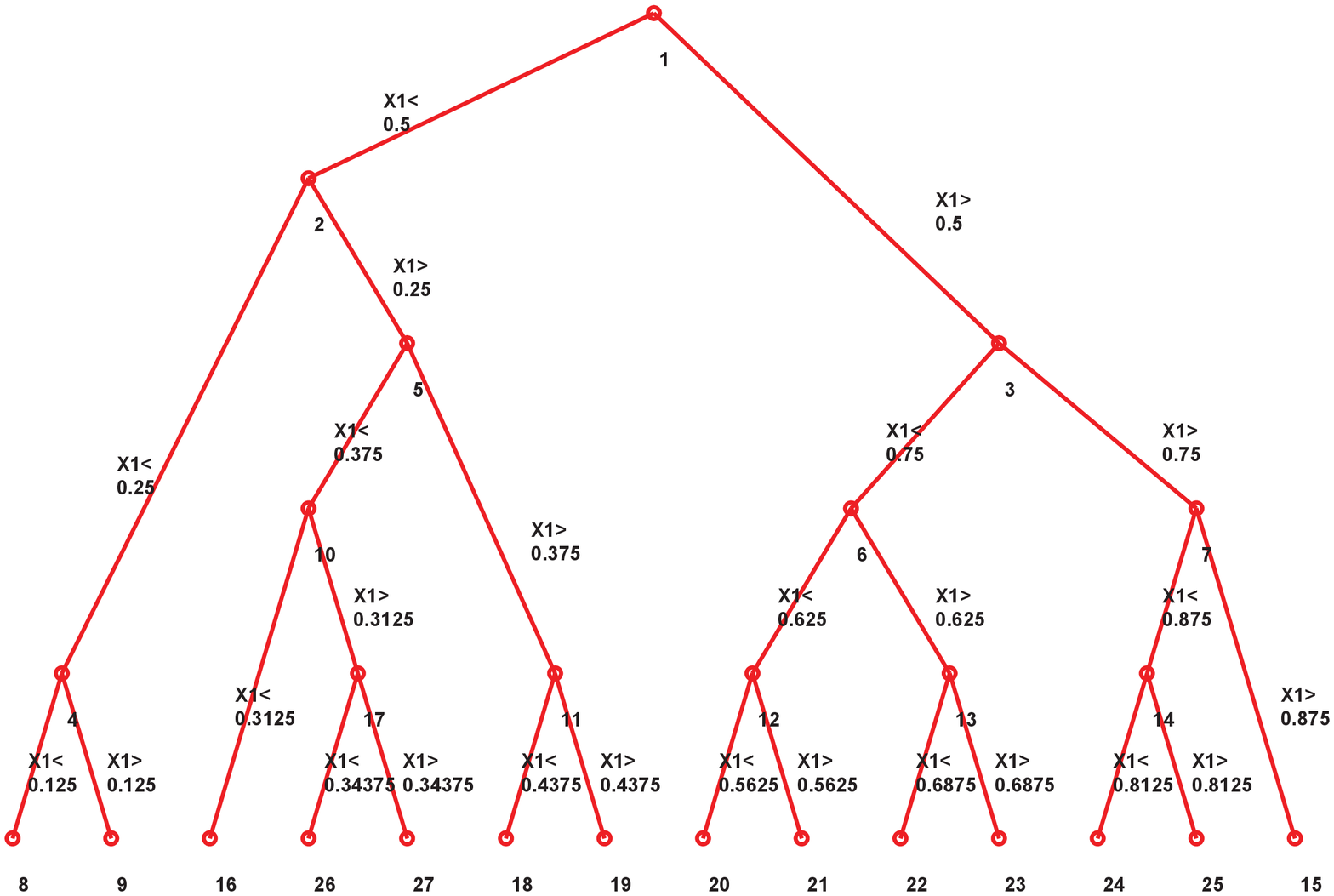} &
  \includegraphics[scale=0.42]{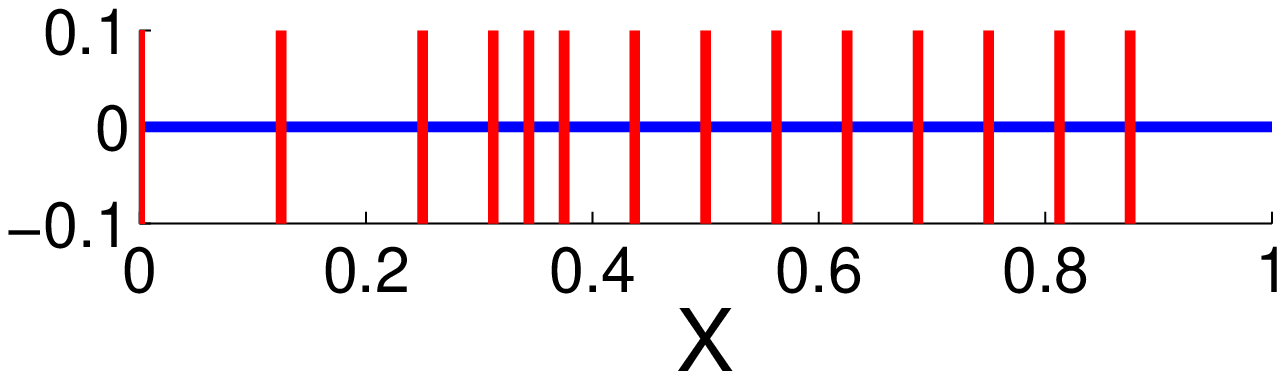}
   \\
   (a) & (b)
   \end{tabular}
  \end{small}
   \caption{(a) Estimated tree structure; (b) corresponding partitions}
   \label{fig:chaintree}
\end{figure}

We generate an equal-sized held-out dataset in the same
manner, using the same $\mu$ and $\Sigma^t$.  Greedy
Go-CART is used to estimate the dyadic tree structure and
corresponding inverse covariance matrices; these are displayed in
Figure \ref{fig:chaintree}.

\subsection{Chain Structure}

\begin{figure}[t]
\centering
  \begin{small}
  \begin{tabular}{cc}
  \includegraphics[scale=0.5]{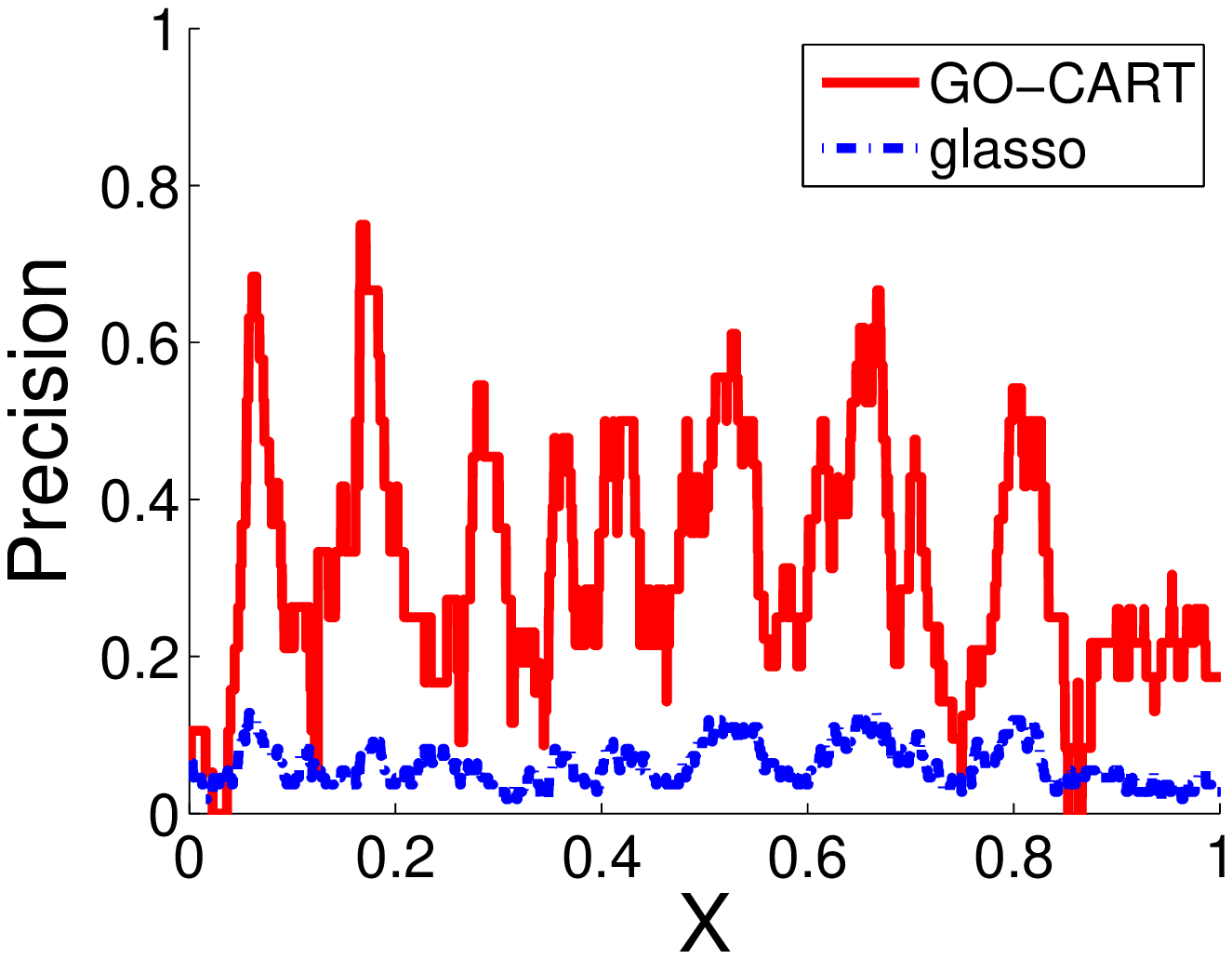} &
  \includegraphics[scale=0.5]{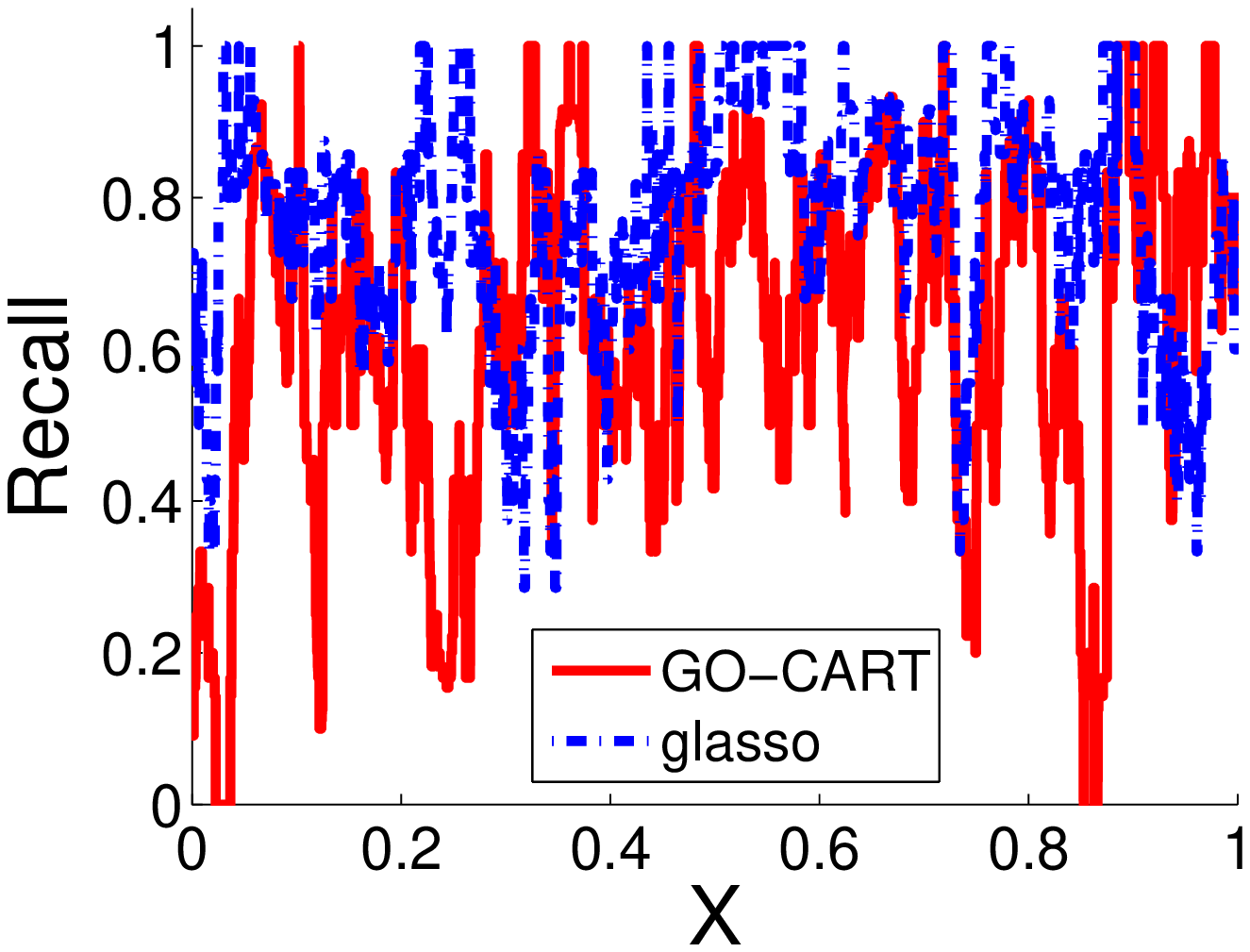} \\
     (a) & (b)\\
  \includegraphics[scale=0.5]{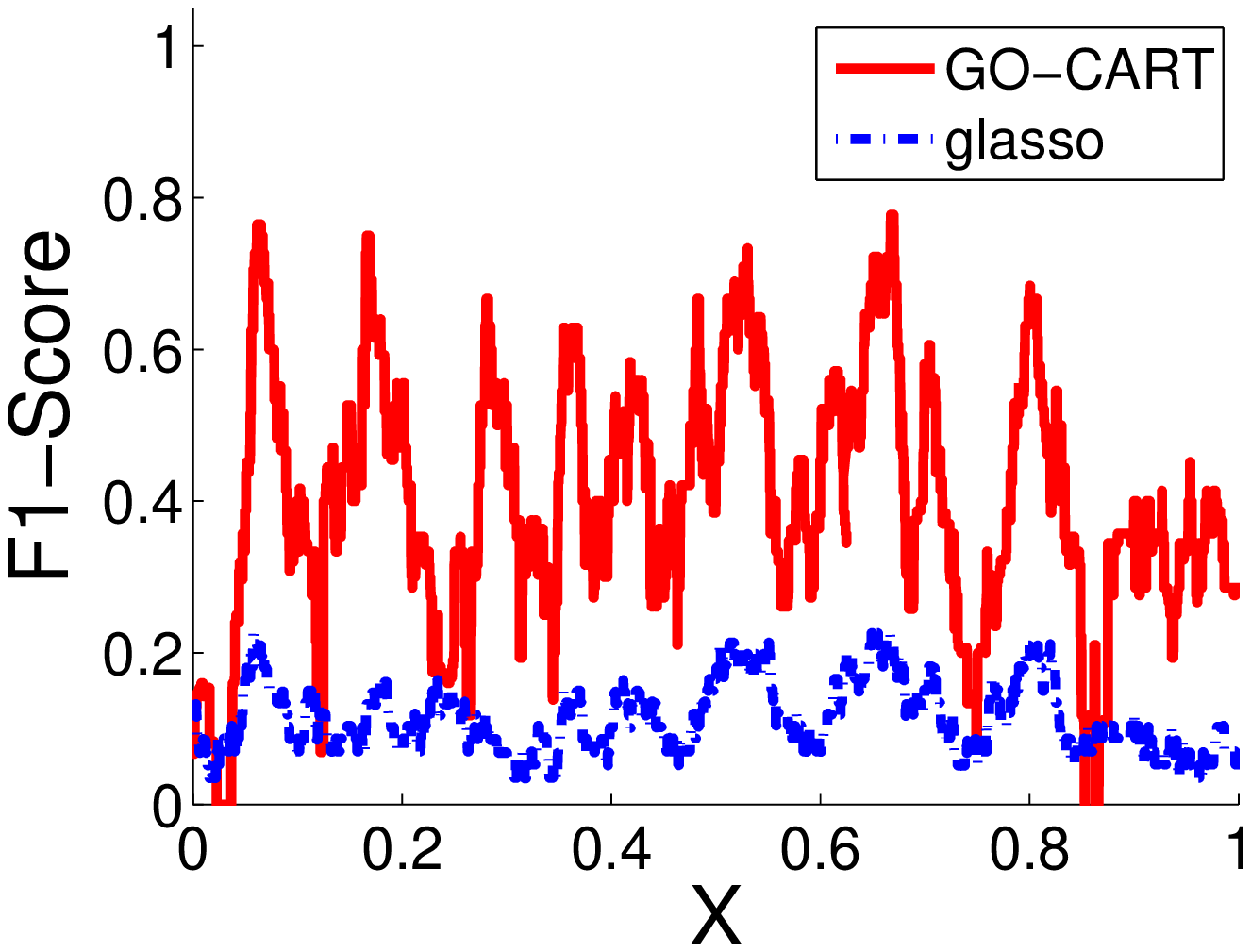}  &
  \includegraphics[scale=0.4]{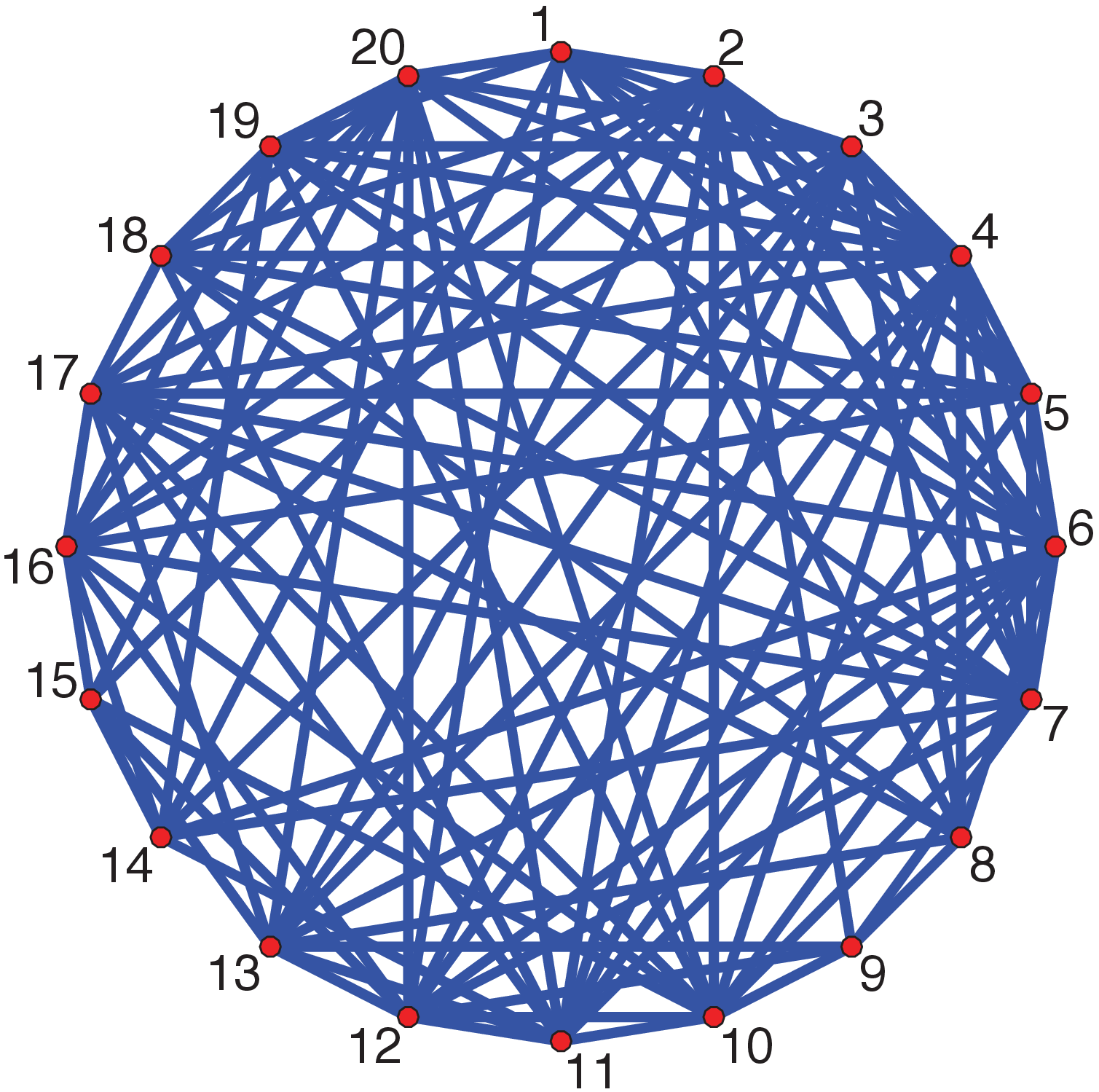}
   \\
  (c) & (d)
   \end{tabular}
  \end{small}
   \caption{Comparison of our algorithm with glasso (a) Precision; (b) Recall; (c) $F_1$-score; (d) Estimated graph by applying glasso on  the entire dataset}
   \label{fig:chainstat}
\end{figure}

To examine the recovery quality of the underlying graph structure,
we compare our estimated graphs to the graph estimated by directly
applying the glasso to the entire dataset. Comparisons in terms of
precision, recall and $F_1$-score are given in Figure \ref{fig:chainstat} (a),
(b) and (c) respectively.  As we can see, the partition-based method achieves much
higher precision and $F_1$-Score. As for recall, glasso is 
slightly better, due to the fact that the glasso graphs estimated on
the entire data are very dense, as shown in \ref{fig:chainstat} (d).
The dense graphs lead to fewer false negatives (thus large recall)
but many false positives (thus small precision).

\subsection{Two-way Grid Structure}

In this section, we apply Go-CART to a two dimensional design
$X$. The underlying graph structures and $Y$ are generated in manner
similar to that used in the previous section.  In particular, we
generate equally spaced $x_1, \ldots,x_n \in \mathbb{R}^{2}$ with
$n=10,000$ on the unit two-dimensional grid $[0,1]^2$. We generate an
Erd\"os-R\'enyi random graph $G^{1,1}=(V^{1,1},E^{1,1})$ with 
$p=20$ vertices, $|E|=10$ edges, and 
maximum node degree four, then construct the graphs for each $x$
along diagonals. More precisely, for each pair of $i,j$, where $1
\leq i \leq 100$ and $1 \leq j \leq 100$, we randomly select either
$G^{i-1,j}$ (if it exists) or $G^{i,j-1}$ (if it exists) with equal
probability as the basis graph. Then, we construct the graph
$G^{i,j}=(V^{i,j}, E^{i,j})$ by removing one edge and adding one
edge with probability $0.05$ based on the selected basis graph,
taking care that the number of edges is between 5 and 15 and the
maximum degree is still four.  Given the underlying graphs, we
generate the covariance matrix and output $Y$ in the same way as in
the last section.

\begin{figure}[!ht]
\centering
  \begin{small}
  \begin{tabular}{cc}
  \includegraphics[scale=0.35]{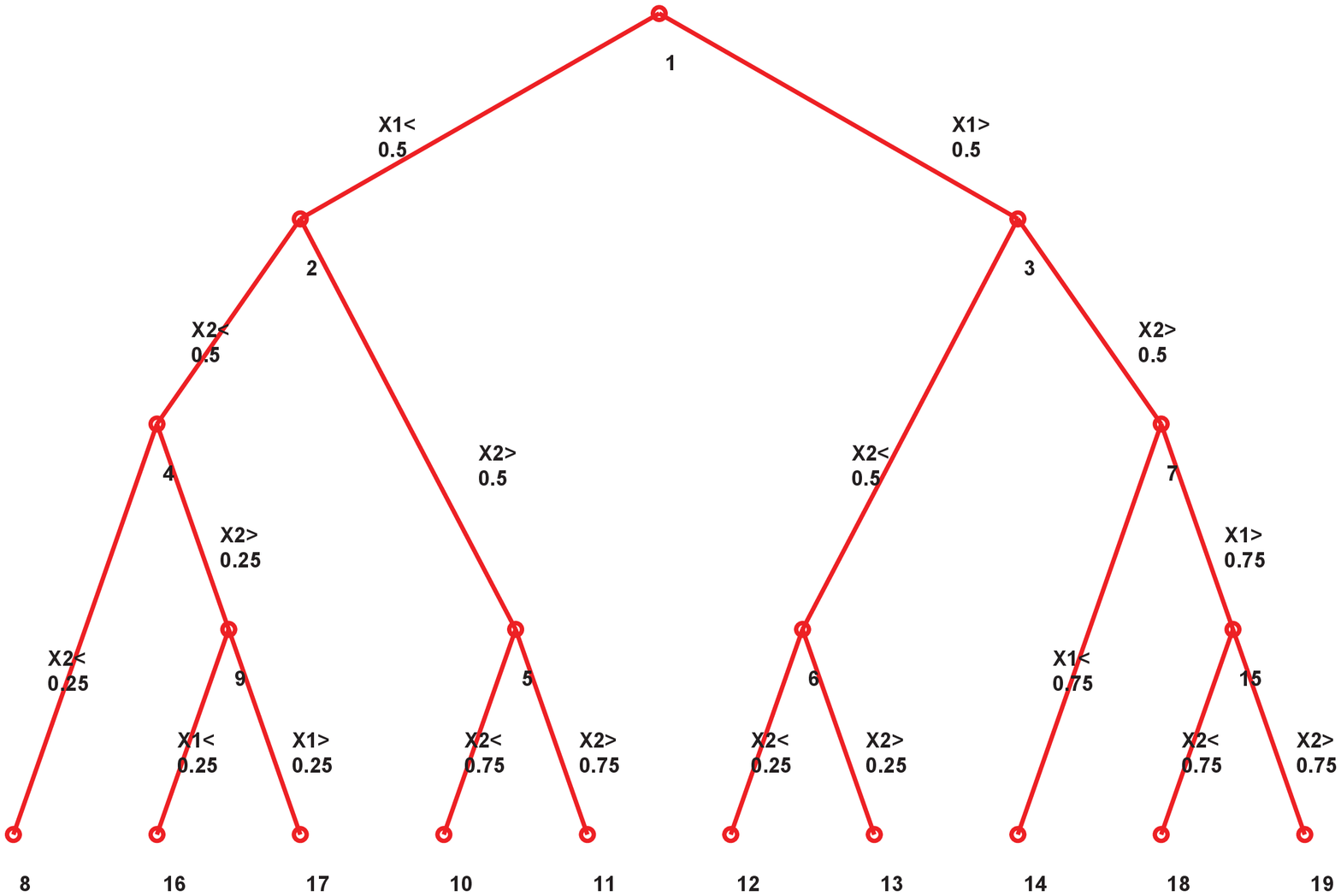} &
  \includegraphics[scale=0.35]{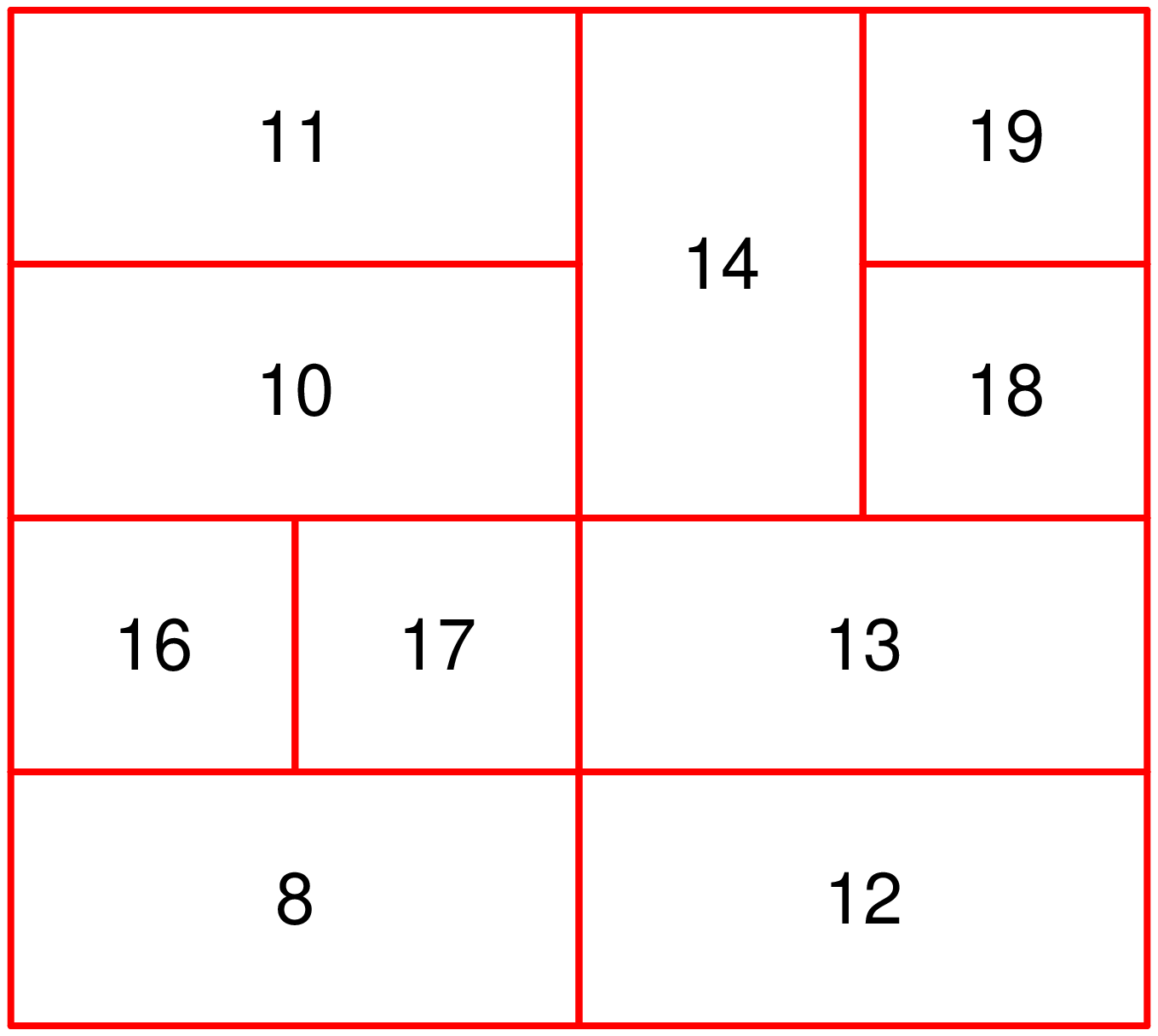}
   \\
   (a) & (b)
   \end{tabular}
  \end{small}
   \caption{(a) Estimated tree structure; (b) estimated partitions where the labels correspond to the index of the leaf node in (a)}
   \label{fig:grid_tree}
\end{figure}

\begin{figure}[!ht]
\centering
  \begin{small}
  \begin{tabular}{cc}
  \includegraphics[scale=0.5]{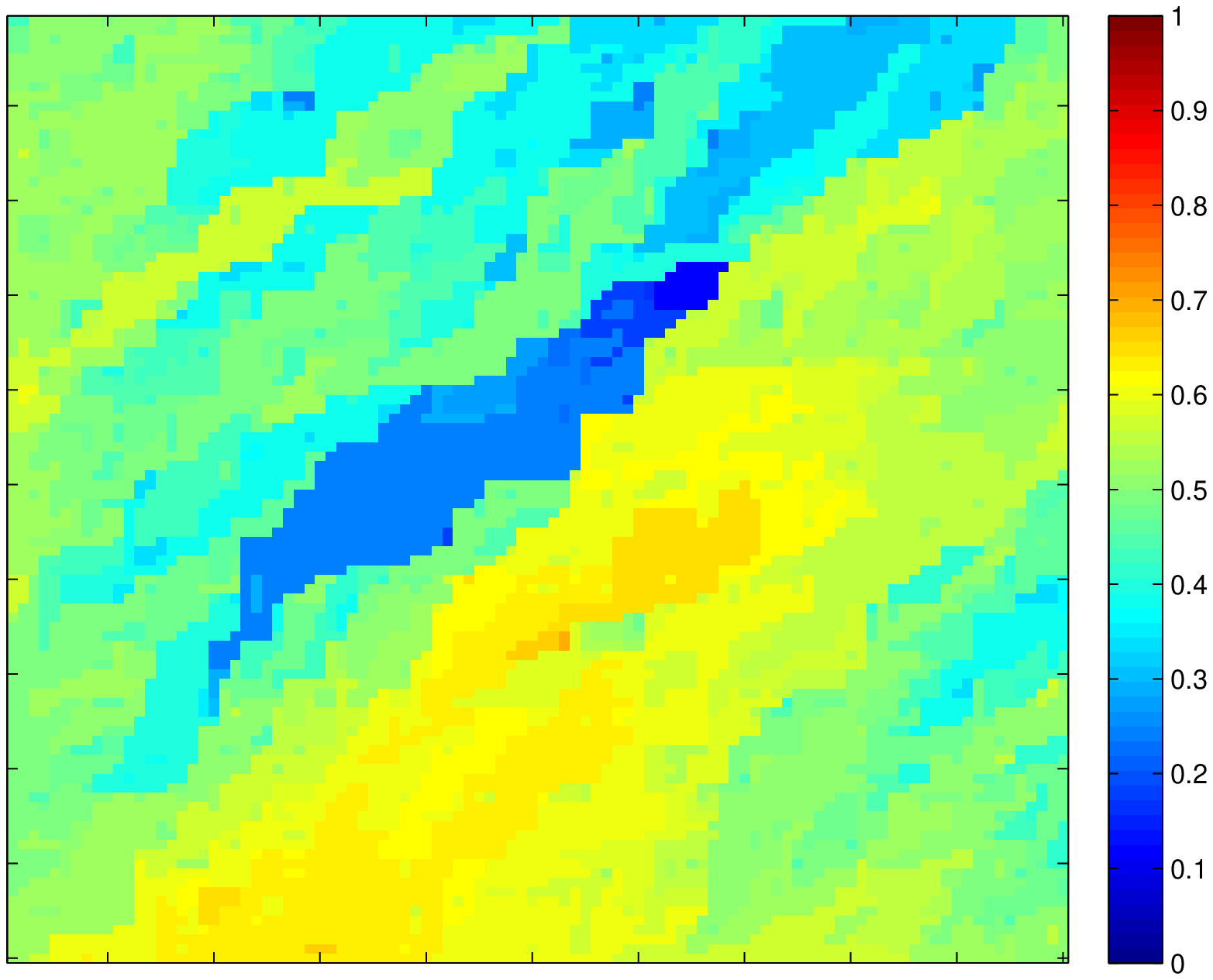} &
  \includegraphics[scale=0.5]{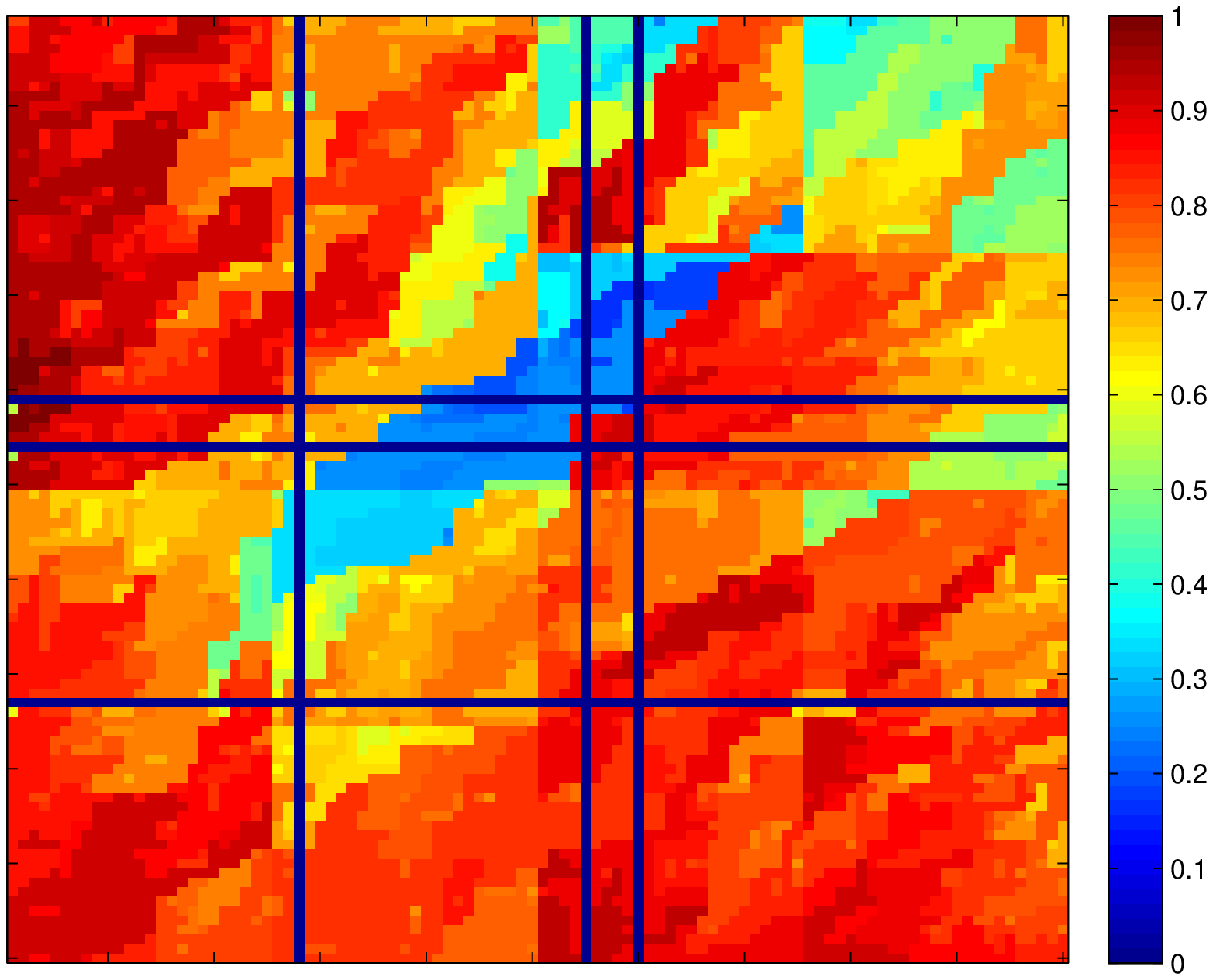}
   \\
   (a) & (b)
   \end{tabular}
  \end{small}
   \caption{(a) Color map of $F_1$-score for glasso run on the entire
     dataset;  (b) color map of $F_1$-score for Go-CART.
   Red indicates large values (approaching 1) and blue
   indicates small values (approaching 0), as shown in the
   color bar.}
   \label{fig:grid_fscore}
\end{figure}

We apply the greedy algorithm to learn the dyadic tree structure and
corresponding inverse covariance matrices, shown in
Figure \ref{fig:grid_tree}. We plot the $F_1$-score obtained by glasso on
the entire data compared against the our method in Figure
\ref{fig:grid_fscore}.  It is seen that for most $x$, the
partitioning method achieves significantly higher $F_1$-score than directly applying
the glasso. Note that since the graphs near the middle part of the
diagonal (the line connecting $[0,1]$ and $[1,0]$) have the greatest
variability, the $F_1$-scores for both methods are
low in this region.

\clearpage

\bibliography{local}

\begin{thebibliography}{15}

\bibitem[\protect\citeauthoryear{Banerjee, Ghaoui and
  d'Aspremont}{2008}]{Banerjee:08}
\begin{barticle}[author]
\bauthor{\bsnm{Banerjee},~\bfnm{Onureena}\binits{O.}},
  \bauthor{\bsnm{Ghaoui},~\bfnm{Laurent~El}\binits{L.~E.}} \AND
  \bauthor{\bsnm{d'Aspremont},~\bfnm{Alexandre}\binits{A.}}
(\byear{2008}).
\btitle{Model selection through sparse maximum likelihood estimation}.
\bjournal{Journal of Machine Learning Research}
\bvolume{9}
\bpages{485--516}.
\end{barticle}
\endbibitem

\bibitem[\protect\citeauthoryear{Blanchard {\it et~al.}}{2007}]{blanchard07}
\begin{barticle}[author]
\bauthor{\bsnm{Blanchard},~\bfnm{G.}\binits{G.}},
  \bauthor{\bsnm{Sch\"{a}fer},~\bfnm{C.}\binits{C.}},
  \bauthor{\bsnm{Rozenholc},~\bfnm{Y.}\binits{Y.}} \AND
  \bauthor{\bsnm{M\"{u}ller},~\bfnm{K.-R.}\binits{K.-R.}}
(\byear{2007}).
\btitle{Optimal dyadic decision trees}.
\bjournal{Mach. Learn.}
\bvolume{66}
\bpages{209--241}.
\end{barticle}
\endbibitem

\bibitem[\protect\citeauthoryear{Breiman {\it et~al.}}{1984}]{cart:84}
\begin{bbook}[author]
\bauthor{\bsnm{Breiman},~\bfnm{Leo}\binits{L.}},
  \bauthor{\bsnm{Friedman},~\bfnm{Jerome}\binits{J.}},
  \bauthor{\bsnm{Stone},~\bfnm{Charles~J.}\binits{C.~J.}} \AND
  \bauthor{\bsnm{Olshen},~\bfnm{R.A.}\binits{R.}}
(\byear{1984}).
\btitle{Classification and regression trees}.
\bpublisher{Wadsworth Publishing Co Inc}.
\end{bbook}
\endbibitem

\bibitem[\protect\citeauthoryear{Edwards}{1995}]{Edwa:1995}
\begin{bbook}[author]
\bauthor{\bsnm{Edwards},~\bfnm{David}\binits{D.}}
(\byear{1995}).
\btitle{Introduction to graphical modelling}.
\bpublisher{Springer-Verlag Inc}.
\end{bbook}
\endbibitem

\bibitem[\protect\citeauthoryear{Friedman, Hastie and
  Tibshirani}{2007}]{FHT:07}
\begin{barticle}[author]
\bauthor{\bsnm{Friedman},~\bfnm{Jerome~H.}\binits{J.~H.}},
  \bauthor{\bsnm{Hastie},~\bfnm{Trevor}\binits{T.}} \AND
  \bauthor{\bsnm{Tibshirani},~\bfnm{Robert}\binits{R.}}
(\byear{2007}).
\btitle{Sparse inverse covariance estimation with the graphical lasso}.
\bjournal{Biostatistics}
\bvolume{9}
\bpages{432--441}.
\end{barticle}
\endbibitem

\bibitem[\protect\citeauthoryear{IPCC}{2007}]{IPCC}
\begin{barticle}[author]
\bauthor{\bsnm{IPCC},~}
(\byear{2007}).
\btitle{\emph{Climate Change 2007--The Physical Science Basis} IPCC Fourth
  Assessment Report}.
\end{barticle}
\endbibitem

\bibitem[\protect\citeauthoryear{Lauritzen}{1996}]{laur:1996}
\begin{bbook}[author]
\bauthor{\bsnm{Lauritzen},~\bfnm{Steffen~L.}\binits{S.~L.}}
(\byear{1996}).
\btitle{Graphical Models}.
\bpublisher{Oxford University Press}.
\end{bbook}
\endbibitem

\bibitem[\protect\citeauthoryear{Liu, Lafferty and Wasserman}{2010}]{Liu:10}
\begin{bunpublished}[author]
\bauthor{\bsnm{Liu},~\bfnm{Han}\binits{H.}},
  \bauthor{\bsnm{Lafferty},~\bfnm{John}\binits{J.}} \AND
  \bauthor{\bsnm{Wasserman},~\bfnm{Larry}\binits{L.}}
(\byear{2010}).
\btitle{Tree Density Estimation}.
\bnote{arXiv:1001.1557v1 [stat.ML] 10 Jan 2010}.
\end{bunpublished}
\endbibitem

\bibitem[\protect\citeauthoryear{Lozano {\it et~al.}}{2009}]{Lozano:09}
\begin{binproceedings}[author]
\bauthor{\bsnm{Lozano},~\bfnm{Aurelie~C.}\binits{A.~C.}},
  \bauthor{\bsnm{Li},~\bfnm{Hongfei}\binits{H.}},
  \bauthor{\bsnm{Niculescu-Mizil},~\bfnm{Alexandru}\binits{A.}},
  \bauthor{\bsnm{Liu},~\bfnm{Yan}\binits{Y.}},
  \bauthor{\bsnm{Perlich},~\bfnm{Claudia}\binits{C.}},
  \bauthor{\bsnm{Hosking},~\bfnm{Jonathan}\binits{J.}} \AND
  \bauthor{\bsnm{Abe},~\bfnm{Naoki}\binits{N.}}
(\byear{2009}).
\btitle{Spatial-temporal causal modeling for climate change attribution}.
In \bbooktitle{ACM SIGKDD}.
\end{binproceedings}
\endbibitem

\bibitem[\protect\citeauthoryear{Ravikumar {\it
  et~al.}}{2009}]{Ravikumar:Gauss:09}
\begin{binproceedings}[author]
\bauthor{\bsnm{Ravikumar},~\bfnm{Pradeep}\binits{P.}},
  \bauthor{\bsnm{Wainwright},~\bfnm{Martin}\binits{M.}},
  \bauthor{\bsnm{Raskutti},~\bfnm{Garvesh}\binits{G.}} \AND
  \bauthor{\bsnm{Yu},~\bfnm{Bin}\binits{B.}}
(\byear{2009}).
\btitle{Model Selection in {G}aussian Graphical Models: {H}igh-Dimensional
  Consistency of $\ell_1$-regularized {MLE}}.
In \bbooktitle{Advances in Neural Information Processing Systems 22}.
\bpublisher{MIT Press}, \baddress{Cambridge, MA}.
\end{binproceedings}
\endbibitem

\bibitem[\protect\citeauthoryear{Rothman {\it et~al.}}{2008}]{Rothman:08}
\begin{barticle}[author]
\bauthor{\bsnm{Rothman},~\bfnm{Adam~J.}\binits{A.~J.}},
  \bauthor{\bsnm{Bickel},~\bfnm{Peter~J.}\binits{P.~J.}},
  \bauthor{\bsnm{Levina},~\bfnm{Elizaveta}\binits{E.}} \AND
  \bauthor{\bsnm{Zhu},~\bfnm{Ji}\binits{J.}}
(\byear{2008}).
\btitle{Sparse permutation invariant covariance estimation}.
\bjournal{Electronic Journal of Statistics}
\bvolume{2}
\bpages{494--515}.
\end{barticle}
\endbibitem

\bibitem[\protect\citeauthoryear{Scott and Nowak}{2006}]{scott06}
\begin{barticle}[author]
\bauthor{\bsnm{Scott},~\bfnm{C.}\binits{C.}} \AND
  \bauthor{\bsnm{Nowak},~\bfnm{R.D.}\binits{R.}}
(\byear{2006}).
\btitle{Minimax-optimal classification with dyadic decision trees}.
\bjournal{Information Theory, IEEE Transactions on}
\bvolume{52}
\bpages{1335-1353}.
\end{barticle}
\endbibitem

\bibitem[\protect\citeauthoryear{Whittaker}{1990}]{whit:1990}
\begin{bbook}[author]
\bauthor{\bsnm{Whittaker},~\bfnm{J.}\binits{J.}}
(\byear{1990}).
\btitle{Graphical Models in Applied Multivariate Statistics}.
\bpublisher{Wiley}.
\end{bbook}
\endbibitem

\bibitem[\protect\citeauthoryear{Yuan and Lin}{2007}]{Yuan:Lin:07}
\begin{barticle}[author]
\bauthor{\bsnm{Yuan},~\bfnm{Ming}\binits{M.}} \AND
  \bauthor{\bsnm{Lin},~\bfnm{Yi}\binits{Y.}}
(\byear{2007}).
\btitle{Model selection and estimation in the {G}aussian graphical model}.
\bjournal{Biometrika}
\bvolume{94}
\bpages{19--35}.
\end{barticle}
\endbibitem

\bibitem[\protect\citeauthoryear{Zhou, Lafferty and Wasserman}{2010}]{Zhou:10}
\begin{barticle}[author]
\bauthor{\bsnm{Zhou},~\bfnm{Shuheng}\binits{S.}},
  \bauthor{\bsnm{Lafferty},~\bfnm{John}\binits{J.}} \AND
  \bauthor{\bsnm{Wasserman},~\bfnm{Larry}\binits{L.}}
(\byear{2010}).
\btitle{Time Varying Undirected Graphs}.
\bjournal{Machine Learning}
\bvolume{78}.
\end{barticle}
\endbibitem

\end{thebibliography}

\end{document}